\documentclass[11pt]{article}

\usepackage[final]{acl}
\usepackage{algorithm}
\usepackage{algpseudocode} 
\usepackage[table]{xcolor}
\usepackage{makecell}
\usepackage{times}
\usepackage{latexsym}
\usepackage{booktabs}
\usepackage{multicol}
\usepackage{multirow}
\usepackage{amsmath,amssymb} 
\usepackage[T1]{fontenc}
\usepackage{enumitem}
\usepackage{orcidlink}

\usepackage[utf8]{inputenc}

\usepackage{microtype}
\usepackage{url}

\usepackage{inconsolata}

\usepackage{graphicx}

\title{VisCache: Visual KV Cache Pruning for Efficient Vision Large Language Model Inference}

\author{
  Lyuke Wang\textsuperscript{1,2,3}\,\orcidlink{0009-0009-8908-7168},
  Zhuo Li\textsuperscript{2,3}\,\thanks{Technical lead}\orcidlink{0009-0000-6451-4877},
  Guangxu Zhu\textsuperscript{1,2,3,4}\,\thanks{Corresponding to zhuguangxu@cuhk.edu.cn}\orcidlink{0000-0001-9532-9201} \\
  \\
  \textsuperscript{1}Shenzhen International Center for Industrial and Applied Mathematics \\
  \textsuperscript{2}Shenzhen Research Institute of Big Data \\
  \textsuperscript{3}The Chinese University of Hong Kong, Shenzhen \\
  \textsuperscript{4}Shenzhen Loop Area Institute \\
}

\begin{document}
\maketitle
\begin{abstract}

While Vision Large Language Models (VLLMs) have achieved remarkable success in multimodal reasoning, their long-context inference remains prohibitively expensive due to the massive computation and memory overhead of visual Key-Value (KV) caches. Existing KV compression methods often apply uniform pruning across visual tokens and layers, leading to substantial information loss and degraded performance.
To address this challenge, we propose \textbf{VisCache}, a plug-and-play framework for coarse-to-fine \textbf{Vis}ual KV \textbf{Cache} pruning without training, which consists of two synergistic stages. First, a lightweight VLM filters temporal redundancy by selectively forwarding semantically informative keyframes. Second, we introduce {PruneKV}, a surgical KV compression algorithm tailored to the attention dynamics of VLLMs. Unlike rigid pruning strategies, PruneKV adopts a parabolic layer-wise budget allocation together with an asymmetric update mechanism that selectively prunes keys while fusing values, thereby preserving critical contextual information. 
Extensive experiments demonstrate that VisCache substantially improves inference efficiency, achieving up to {2.35$\times$ speedup} and significant memory reduction while maintaining competitive performance with only {19--28\%} KV cache retention. VisCache consistently outperforms existing baselines, establishing a new Pareto frontier between efficiency and performance for long-context VLLM inference. Code is available at \url{https://github.com/Wlklk/VisCache}
\end{abstract}

\section{Introduction}

The integration of vision into Large Language Models (VLLMs) has unlocked a new frontier in artificial intelligence~\cite{bai2023qwen,chen2024internvl,li2024llava,liu2023visual,liu2024improved,wu2024deepseek}.
By bridging the modality gap between textual understanding and visual perception, models such as LLaVA~\cite{liu2023visual} and GPT-4V~\cite{yang2023dawn} have demonstrated remarkable proficiency in complex tasks ranging from long-video understanding to real-time visual conversational scenarios and derivative tasks~\cite{tao2025dycoke,zhang2024sparsevlm,yang2025visionzip,liu2025laco,hua2025v2xum,varma2025retracted,hu-etal-2022-graph,10.1145/3746027.3755726,dai-etal-2026-psyche,yuan2023explainablefinegrained3dgrounding,hu2026agentmental,10904873,li-etal-2025-add}.
While VLLMs have been predominantly deployed in visual-centric tasks such as Video Summarization (VS)~\cite{bai2025qwen3,maaz2024video} and Visual Question Answering (VQA)~\cite{li2022blip,wang2025internvl3_5,zhu2025internvl3}, their practical adoption is severely hindered by substantial latency bottlenecks and massive memory requirements~\cite{yang2025topv,ye2025voco,li2025madakv}, particularly when processing longer videos and higher-resolution streams.

Specifically, long videos produce massive visual tokens that dramatically extend the input length fed into the LLM backbone, creating two interrelated bottlenecks. First, the Key-Value (KV) cache, inflated by the sheer volume of visual tokens, dominates GPU memory consumption and intensifies memory bandwidth contention, severely constraining throughput during long-context inference~\cite{wan2025meda,pope2023efficiently,hooper2024kvquant}. Second, attention computation scales quadratically with sequence length, further amplifying latency overhead through increasingly expensive matrix multiplications~\cite{liu2025zsmerge,shao2025holitom}. Together, these intertwined storage and computational pressures fundamentally limit the scalability of VLLMs for long-video understanding.


Naively discarding visual tokens reduces computational cost but often incurs severe performance degradation~\cite{tan2025tokencarve}, motivating more principled KV cache compression methods~\cite{zhang2023h2o,li2024snapkv,ainslie2023gqa,kwon2023efficient}.
Among them, quantization~\cite{liu2024kivi,lin2024duquant,sun2024flatquant,sheng2023flexgen} and low-rank decomposition~\cite{chang2024palu,saxena2024eigen,sun2024shadowkv} are widely adopted to reduce memory and computation by lowering precision or exploiting structural redundancy.
However, quantization is vulnerable to activation outliers~\cite{ashkboos2024quarot,xiao2023smoothquant}, and aggressive rank constraints can impair attention capacity~\cite{chang2024palu,saxena2024eigen,sun2024shadowkv}.
More fundamentally, both rely on coarse compression schemes that overlook the intricate information flow within the model.
Recent studies further reveal that visual redundancy in VLLMs is highly structured rather than uniformly distributed: only a subset of video frames is relevant to a given query~\cite{lee2025refocus,zhou2025reason}, different transformer layers contribute unevenly to downstream reasoning~\cite{zhang2025cross,wang2024prefixkv}, and keys and values serve fundamentally different roles in attention computation~\cite{vaswani2017attention}. These observations indicate that effective visual KV cache compression should jointly consider temporal relevance, layer-wise importance, and the asymmetric roles of keys and values.

\begin{figure}[!t]
    \centering
    \includegraphics[width=0.48\textwidth,trim=0 0 0 0, clip]{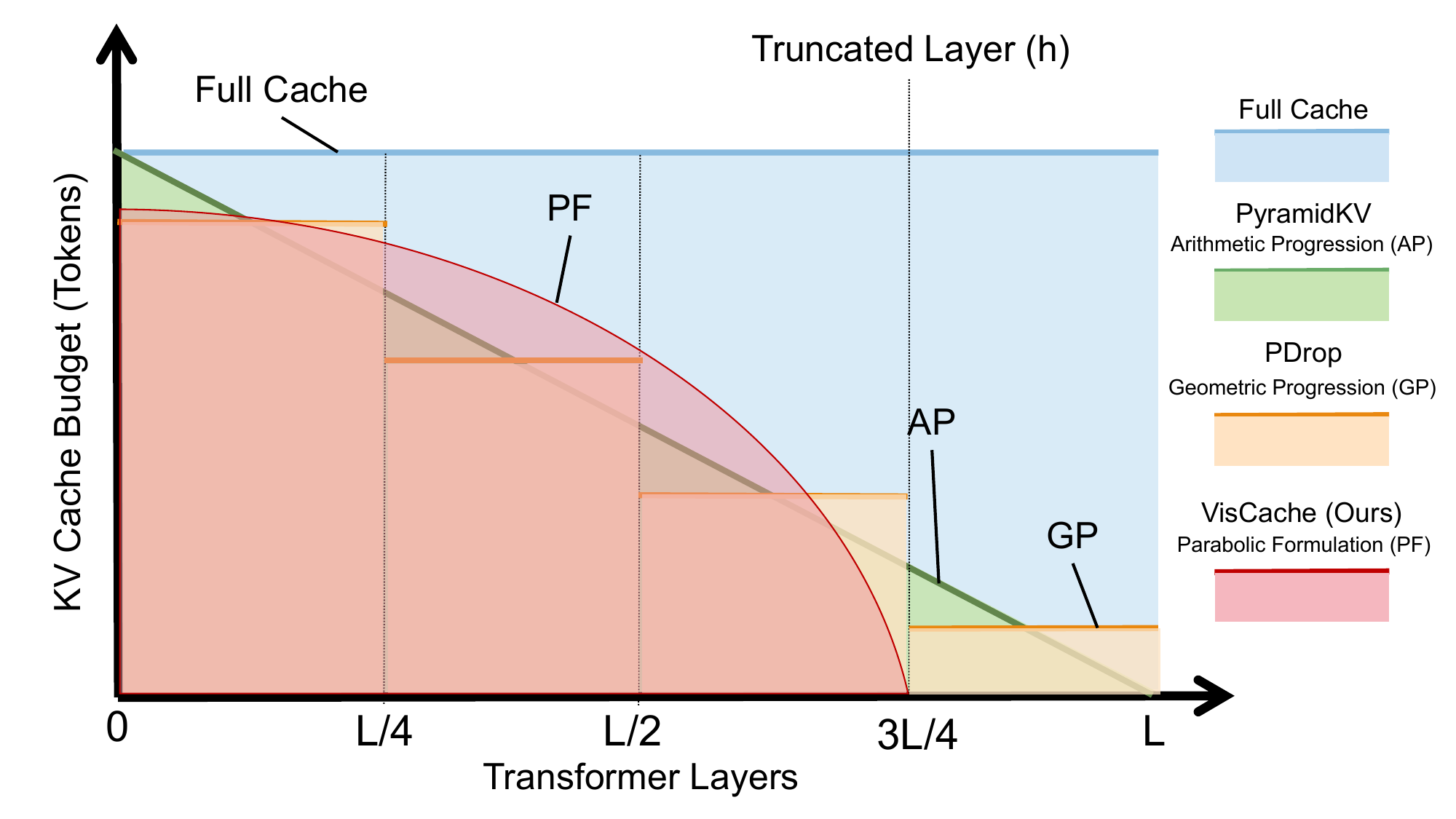}
    \caption{
        Visualization of different plug-and-play layer-wise KV cache compression methods. 
        It illustrates that VisCache differs fundamentally from other baseline approaches in terms of layer-wise budgets. 
    }
    \label{hierarchical_compress_comparison}\vspace{-1.5em}
\end{figure}

In this paper, we propose \textbf{VisCache}, a training-free and plug-and-play framework for coarse-to-fine \textbf{Vis}ual KV \textbf{Cache} pruning, which reduces redundancy at two complementary stages: semantic filtering before inference and structural compression during inference. At the input level, VisCache performs prompt-aware temporal filtering to eliminate redundant frames before they enter the VLLM. Concretely, a lightweight Vision-Language ``scout'' model (e.g., CLIP~\citep{radford2021learning}) identifies task-relevant keyframes guided by prompt-aware reasoning and the Maximal Marginal Relevance (MMR) principle~\cite{li2016multimedia}, selecting a compact yet diverse subset of frames that preserves semantic coverage while substantially reducing visual token redundancy.

At the model level, we introduce PruneKV, a layer-aware visual KV compression algorithm. Instead of applying uniform pruning across layers and token types, PruneKV dynamically allocates layer-wise compression budgets following a parabolic schedule. As shown in Figure~\ref{hierarchical_compress_comparison}, PruneKV preserves more visual tokens in early layers that encode fine-grained spatial details and progressively fewer in deeper layers that capture abstract semantics, with visual KV entries beyond a truncation threshold entirely evicted. Moreover, PruneKV adopts an asymmetric update strategy that treats keys and values differently: unimportant keys are selectively discarded, while their corresponding values are fused through weighted aggregation into the retained tokens to preserve contextual information. This attention-aware design substantially reduces KV cache redundancy while maintaining stable reasoning performance. The two stages of VisCache operate synergistically: the scout reduces redundant visual inputs before inference, while PruneKV further compresses the internal KV cache during generation.
Extensive experiments demonstrate that VisCache achieves up to $2.35\times$ inference speedup while retaining only 19--28\% of the KV cache, consistently outperforming existing compression baselines across multiple video understanding benchmarks. Comprehensive ablation studies suggest that each proposed module is essential to VisCache's strong performance.
We summarize our contributions as follows:
\begin{itemize}[leftmargin=*,parsep=-0.3em]
    \item We propose {VisCache}, a training-free and plug-and-play coarse-to-fine framework that jointly performs prompt-aware frame filtering and visual KV cache compression for efficient long-context VLLM inference.
    
    \item We introduce {PruneKV}, a layer-aware KV compression algorithm with parabolic budget allocation and asymmetric key-value updates that better align pruning with attention dynamics.
    
    \item Extensive experiments on long-video understanding benchmarks show that VisCache consistently achieves superior efficiency-performance trade-offs over existing baselines under aggressive KV cache compression.
\end{itemize}

\section{Preliminary}
\paragraph{VLLM Prefilling.} Given a video input consisting of $M_V$ frames, the visual encoder of the VLLM sequentially processes each frame $f$ that contains $N_V$ visual tokens and projects them into a shared embedding space of dimension $d$. 
As a result, the visual representations of all frames are aggregated into a visual embedding matrix $\mathbf{H}_v \in \mathbb{R}^{M_V N_V \times d}$.
In parallel, the corresponding textual prompt $T = \{x_i\}_{i=1}^{N_T}$ is fed into the text embedding layer, producing a text embedding matrix $\mathbf{H}_q \in \mathbb{R}^{N_T \times d}$, where $x_i$ denotes the $i$-th input textual token.
The visual and textual embeddings are then concatenated along the token dimension to form the unified input 
$\mathbf{H} = \operatorname{concat}\!\left[\mathbf{H}_v,\ \mathbf{H}_q\right] \in \mathbb{R}^{(M_V N_V + N_T) \times d}.$ For a VLLM composed of $L$ transformer layers, the self-attention module at each layer $l \in \{1, \dots, L\}$ is parameterized by three projection matrices,
$\mathbf{W}_Q^{l}, \mathbf{W}_K^{l}, \mathbf{W}_V^{l} \in \mathbb{R}^{d\times d}$.
These matrices are used to compute the query, key, and value representations as
\begin{equation}
    \vspace{-0.2em}
\small
\mathbf{Q}^{l} = \mathbf{H}\mathbf{W}_Q^{l},\quad
\mathbf{K}^{l} = \mathbf{H}\mathbf{W}_K^{l},\quad 
\mathbf{V}^{l} = \mathbf{H}\mathbf{W}_V^{l}, \nonumber
    \vspace{-0.2em}
\end{equation}
where $\mathbf{K}^{l}$ and $\mathbf{V}^{l}$ are cached and reused during the subsequent decoding stage.

\paragraph{VLLM Decoding.} During decoding, the VLLM generates output tokens in an autoregressive (AR) manner by reusing the KV cache constructed during the prefilling stage.
At decoding step $t$, the embedding of the previously generated token is denoted as $\mathbf{h}_t \in \mathbb{R}^{1 \times d}$.
The KV cache is then incrementally updated as:
\begin{equation}
    \vspace{-0.2em}
\small
\begin{aligned}
\mathbf{K}^{l}\leftarrow \operatorname{concat}\!\left[\mathbf{K}^{l},\ \mathbf{h}_t \mathbf{W}_K^{l}\right],
\mathbf{V}^{l}\leftarrow \operatorname{concat}\!\left[\mathbf{V}^{l},\ \mathbf{h}_t \mathbf{W}_V^{l}\right], \nonumber
\end{aligned}
    \vspace{-0.2em}
\end{equation}
which are then used together with the current query vector to compute the attention output and in turn produce the next token $y_{t+1}$.
This process repeats until a termination condition is met or the maximum generation length is reached.

\begin{figure*}[!t]
    \centering
    \includegraphics[width=1.0\textwidth, trim=0 0 0 0, clip ]{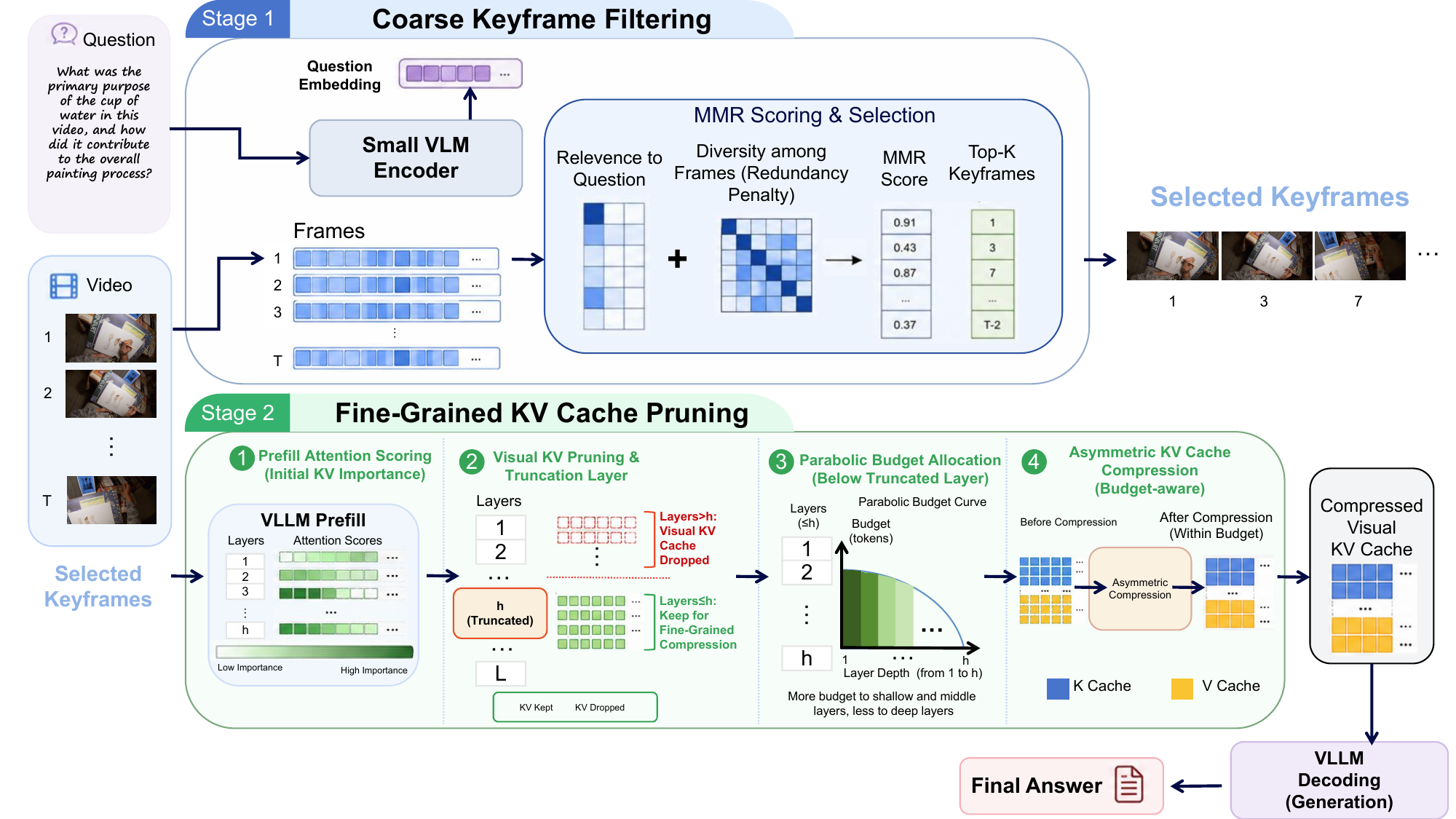}
    \caption{
    {Overview of VisCache.}
    The framework operates in two synergistic stages: 
    (I) A lightweight scout VLM performs prompt-aware keyframe filtering via MMR to eliminate temporal redundancy at the source; 
    (II) During inference, the VLLM constructs the KV cache, which is subsequently refined by PruneKV through attention-aware, layer-wise compression.
    } 
    \label{VisCol}
\end{figure*}

\section{Method}

We propose \textbf{VisCache}, a training-free and plug-and-play framework that compresses the visual KV cache through a coarse-to-fine, dual-stage paradigm.
As illustrated in Figure~\ref{VisCol}, VisCache is motivated by two forms of visual redundancy in long-video VLLM inference:
\textit{temporal redundancy} among frames at the input level, and \textit{structural redundancy} in the KV cache across tokens and layers during generation.
Accordingly, VisCache operates in two complementary stages:
a prompt-aware temporal filtering stage that prunes redundant frames before inference, and a layer-aware KV compression stage that surgically reduces the visual KV cache during generation.

\subsection{Prompt-Aware Scout for Temporal Redundancy Filtering}
Given a video with $M_V$ frames and a textual prompt $T$, we employ a lightweight vision-language model (e.g., CLIP~\cite{radford2021learning}) as a \textit{scout} to extract aligned cross-modal representations.
The text encoder $\mathrm{Enc}_{\text{text}}$ and visual encoder $\mathrm{Enc}_{\text{vis}}$ map the prompt $T$ and each frame $f$ into a shared embedding space by $    \mathbf{h}_t = \mathrm{Enc}_{\text{text}}(T),
    \mathbf{h}_f = \mathrm{Enc}_{\text{vis}}(f),
$ 
where $\mathbf{h}_t \in \mathbb{R}^d$ and $\mathbf{h}_f \in \mathbb{R}^d$ denote the prompt embedding and the $f$-th frame embedding, respectively.
To select a compact yet diverse subset of keyframes, we adopt the Maximal Marginal Relevance (MMR) criterion~\cite{carbonell1998use}, which balances {prompt relevance} against {inter-frame redundancy}.
Let $\Omega$ denote the set of selected frames. We can have MMR score of a candidate frame $f$:
\begin{equation}
    \vspace{-0.2em}
\lambda \cdot \mathrm{sim}\!\left(\mathbf{h}_f, \mathbf{h}_t\right)
      - (1-\lambda) \cdot \max_{f' \in \Omega}\ \mathrm{sim}\!\left(\mathbf{h}_f, \mathbf{h}_{f'}\right),
    \label{eq:mmr}
    \vspace{-0.2em}
\end{equation}
where $\mathrm{sim}(\cdot,\cdot)$ denotes cosine similarity, and $\lambda \in [0, 1]$ is a hyperparameter controlling the trade-off between relevance and diversity.
The frame with the highest MMR score is iteratively added to $\Omega$ until a target retention ratio (RR) $p$ is reached.
The final set $\Omega$ forms a compact, query-relevant, and minimally redundant keyframe sequence that serves as the visual input to the subsequent VLLM inference.

\subsection{PruneKV: Layer-Aware Visual KV Cache Compression}
\label{subsec:prunekv}
Rather than treating all visual KV entries uniformly, PruneKV is designed around two structural properties of transformer attention: layer-wise heterogeneity in visual token importance and functional asymmetry between keys and values. These observations motivate two components of PruneKV:  layer-aware budget allocation and asymmetric key-value compression. Figure~\ref{hierarchical_compress_comparison} provides an overview of the PruneKV pipeline.

\paragraph{Token Scoring and Parabolic Budget Allocation}

During prefilling, the attention weights from all layers serve as a natural signal for token importance.
For each visual token at position $v$, we aggregate its attention received across all queries and layers:
\begin{equation}
    \vspace{-0.2em}
    s_v = \frac{1}{L} \sum_{l=1}^{L} \sum_{i=1}^{N} A^l_{i,v},
    \label{eq:token_score}
    \vspace{-0.2em}
\end{equation}
where $A^l_{i,v}$ is the attention weight from token $i$ to token $v$ at layer $l$, and $N$ is the total sequence length.
Based on $s_v$ as the importance score, the top $q$ of visual KV cache that VLLM focuses on are selected.

Beyond token-level pruning, we further compress the visual KV cache along the layer dimension.
Visual representations become increasingly abstract with depth~\cite{ghiasi2022visiontransformerslearnvisual,zhang2025cross,wang2024prefixkv}, and the number of truly informative visual tokens decreases accordingly.
We therefore allocate compression budgets in a {parabolic decay}: larger budgets in early layers, tapering to smaller budgets in deeper layers.
Let $h \in (1, L]$ be a truncation threshold such that visual KV entries in layers $l > h$ are fully evicted.
For the remaining layers $l \in [1, h]$, the budget proportion is:
\begin{equation}
    \vspace{-0.2em}
    b_l = 1 - \frac{(l-1)^2}{2(h-1)^2},
    \label{eq:parabolic}
    \vspace{-0.2em}
\end{equation}
which satisfies $b_1 = 1$ and $b_h = 0.5$, decaying slowly in early layers and more steeply in deeper layers. 
Compared with linear~\cite{cai2024pyramidkv} or geometric decay~\cite{xing2024pyramiddrop}, Equation~\eqref{eq:parabolic} preserves more tokens where fine-grained visual details are encoded and compresses more aggressively where representations are sufficiently abstract to tolerate heavy pruning. 
We normalize $\{b_l\}_{l=1}^h$ to satisfy a global retention constraint $\sum_{l=1}^{h} b_l = h \cdot m$, where $m \in (0, 1]$ is the target retention ratio for parabolic budget allocation module.
The final budget for each layer determines how many top-scoring visual tokens are retained in $\mathcal{C}_k$ (the keep set), with the remainder assigned to $\mathcal{C}_d$ (the drop set).

\paragraph{Asymmetric Key-Value Update.}
Existing KV compression methods~\citep{cai2024pyramidkv,xing2024pyramiddrop} typically prune keys and values identically, ignoring their distinct roles in attention. However, the standard scaled dot-product attention reveals a clear functional asymmetry: attention weights are determined solely by $\mathbf{Q}\mathbf{K}^{\top}$, making {keys the relevance selectors} that decide which tokens are attended to, while {values are the information carriers} that encode the content ultimately aggregated into the output.
Simply discarding value vectors therefore risks irreversibly losing contextual information that the dropped tokens originally carried.

We exploit this asymmetry through a {prune-keys, fuse-values} strategy.
For the drop set $\mathcal{C}_d$, key vectors are entirely removed, as their contribution to the attention distribution is negligible by construction (these tokens scored lowest in the importance ranking).
The corresponding value vectors, however, are not discarded but \textit{redistributed} to the retained tokens via similarity-based weighted aggregation.
Concretely, let $\mathbf{V}_k \in \mathbb{R}^{b_l \times d}$ and $\mathbf{V}_d \in \mathbb{R}^{(n-b_l)\times d}$ denote the value caches of the keep set $\mathcal{C}_k$ and drop set $\mathcal{C}_d$, respectively.
We compute a redistribution matrix $\mathbf{\Phi}$ that captures the semantic affinity between the two sets:
\begin{equation}
    \mathbf{\Phi} = \mathrm{Softmax}\!\left( \frac{\mathbf{V}_k \mathbf{V}_d^{\top}}{\tau} \right) \in \mathbb{R}^{b_l \times (n-b_l)},
    \label{eq:redistribution}
\end{equation}
where $\tau$ is a temperature that controls the sharpness of the redistribution (e.g., lower $\tau$ concentrates the dropped values onto fewer retained tokens).
Each row of $\mathbf{\Phi}$ specifies how the $b_l$ retained tokens absorb information from the $n-b_l$ dropped tokens.
The aggregated dropped values, $\mathbf{\Phi} \mathbf{V}_d$, are then fused with the original kept values through a weighted combination:
\begin{equation}
    \mathbf{V}_k^{\text{new}} = \mu \mathbf{V}_k + (1-\mu) \left(\mathbf{\Phi} \mathbf{V}_d\right),
    \label{eq:weight_fusion}
\end{equation}
where $\mu \in [0, 1]$ balances the contribution of the original and redistributed information ($\mu = 1$ recovers standard pruning; $\mu = 0$ replaces kept values entirely with fused dropped values).
The compressed KV pair $(\mathbf{K}_k^{\text{new}}, \mathbf{V}_k^{\text{new}})$ then replaces the original visual cache for all subsequent decoding steps.
A complete description of the PruneKV algorithm is provided in Appendix~\ref{app:algorithm}.

\section{Experiment}

\paragraph{Baselines.}
We mainly compare VisCache with plug-and-play baselines, including: PyramidKV~\cite{cai2024pyramidkv} reduces the KV cache budget layer by layer following an arithmetic progression, forming a pyramidal structure based on attention scores. FastV~\cite{chen2024image} leverages the sparsity of visual attention to retain KV entries selectively. PDrop~\cite{xing2024pyramiddrop} partitions layers into stages and progressively decreases token counts in a geometric manner, enabling stage-wise KV cache compression. Q-Frame~\cite{zhang2025q} performs adaptive frame selection and multi-resolution scales tailored to video understanding.

\begin{table*}[!t]
    \centering
    \caption{{Comparison of different KV cache compression methods on various VQA and VS datasets.} 
    Bold and underlined numbers indicate the best and second-best results, respectively.
    }
    \resizebox{\linewidth}{!}{
      \begin{tabular}{l|c|c|c|cc|cccc}
        \toprule  
        \multirow{2}{*}{Method} & \multirow{2}{*}{RR}   & FLOPs  & FLOPs  & ActCap & DREAM1K & NExTQA  & ActQA & EgoSchema & Avg. \\
          &  & (T) & Ratio & ROUGE-L & ROUGE-L & Acc & Acc & Acc & Acc \\
        \midrule
        \multicolumn{10}{c}{{Qwen2.5-VL-3B-Instruct}} \\
        \midrule    
        Full Cache & 100\% & 14.80 & 100\% & 2.63 & 9.19 & 34.69 & 40.58 & 57.20 & 44.16 \\
        Q-Frame & 40\% & 6.26 & 42\% & 2.42 & 5.86 & 39.52 & 39.42 & 46.40 & 41.78\\
        PyramidKV & 40\% & 1.99 & 13\% & 2.43 & 8.48 & 39.53 &  39.56 & \underline{54.80} & 44.63 \\
        FastV & 40\% & 3.11 & 21\% & 2.43 & 8.63 & 37.67 & \underline{40.65} & 50.40 &42.91 \\
        PDrop & 40\% & 2.99 & 20\% & \textbf{2.46} & 8.63 & 34.24 & 40.21 & 50.00 & 41.48 \\
        \rowcolor{blue!10}
        VisCache $(p=0.75, q=0.95)$ & 40\% & 1.32 & 9\% & 2.35 & \textbf{9.79} & 37.62 & 37.32 & 52.00 & 42.31 \\
        \rowcolor{blue!10}
        VisCache $(p=0.75, q=0.67)$ & 28\% & 1.05 & 7\% & \underline{2.45} & \underline{8.70} & \underline{41.25} & \textbf{40.66} & \textbf{55.00} & \textbf{45.64}\\
        \rowcolor{blue!10}
        VisCache $(p=0.50, q=0.67)$ & 19\% & 0.88 & 6\% & 2.39 & 8.26 & \textbf{41.44} & 38.51 & 54.60 & \underline{44.85} \\
        \midrule  
        \multicolumn{9}{c}{{Qwen2.5-VL-32B-Instruct}} \\
        \midrule
        Full Cache & 100\% & 93.08 & 100\% & 2.82 & 7.87 & 60.71 & 46.47 & 65.20 & 57.46 \\
        Q-Frame & 40\% & 40.80 & 44\% & \underline{2.81} & \textbf{7.84} & 50.08 & \underline{43.86} & 54.60 & 49.51 \\
        PyramidKV & 40\% & 13.88 & 15\% & 2.78 & \underline{7.55} & \underline{61.58}  & 42.96 & \underline{65.20} & \underline{56.58} \\
        FastV & 40\% & 51.63 & 55\% & \underline{2.81} & 7.47 & 60.29 &  38.41 & 61.60 & 53.43 \\
        PDrop & 40\% & 31.03 & 33\% & 2.75 & 7.38 & 61.00 & 41.59 & 63.20 & 55.26 \\
        \rowcolor{blue!10}
        VisCache $(p=0.75, q=0.95)$  & 40\% & 13.56 & 15\% & \textbf{3.23} & 7.32 & 55.74 & \textbf{45.46} & 58.80 & 53.00 \\  
        \rowcolor{blue!10}
        VisCache $(p=0.75, q=0.67)$ & 28\% & 10.75 & 12\% & 2.72 & 7.29 & \textbf{62.08} &  42.69 & \textbf{65.80} & \textbf{56.86} \\
        \rowcolor{blue!10}
        VisCache $(p=0.50, q=0.67)$ & 19\% & 9.08 & 10\% & \underline{2.81} & 7.35 & 56.61 &  41.86 & 65.00 & 54.16 \\
        \bottomrule
      \end{tabular}
    }
    \label{tab:main_result1}
\end{table*}

\paragraph{Benchmarks and Evaluation metrics.}
We evaluate on both video summarization (VS) and visual question answering (VQA) benchmarks.
For VS, we use ActivityNet Captions (ActCap)~\cite{caba2015activitynet} (20K videos, 100K captions) and DREAM1K~\cite{wang2024tarsier} (1,000 clips with dense event descriptions), reporting ROUGE-L~\cite{lin2004rouge} as the generation metric.
For VQA, we consider NExTQA~\cite{xiao2021next} for temporal reasoning, ActivityNet-QA (ActQA)~\cite{yu2019activitynet} (5.8K videos), and EgoSchema~\cite{mangalam2023egoschema} for long‑form comprehension.
We further evaluate on MVBench~\cite{li2024mvbench}, a multi‑task benchmark covering 20 reasoning tasks with multiple‑choice QA pairs.

\paragraph{Implementation Details.}
We implement VisCache on the Qwen2.5-VL series using PyTorch and 4 NVIDIA A100 GPUs (80GB)~\cite{bai2025qwen2}. 
For keyframe selection, we adopt the pretrained CLIP ViT-B/32~\cite{radford2021learning} as the scout model.
The frame-level pruning ratio $p$ (Stage 1) and the token-level retention ratio $q$ (Stage 2) serve as the primary control variables.
Other hyperparameters are fixed as follows: $\lambda = 0.7$ in Equation~\eqref{eq:mmr}, layer truncation threshold $h = \frac{3}{4}L$, average layer-wise budget $m = 0.75$, temperature $\tau = 1.0$ in Equation~\eqref{eq:redistribution}, and fusion weight $\mu = 0.7$ in Equation~\eqref{eq:weight_fusion}.
Throughout the paper, \textit{Retention Ratio} (RR) denotes the fraction of visual KV cache preserved relative to the full cache. 
For VisCache, the overall RR is determined by four factors: the frame filtering ratio $p$, the token-level pruning ratio $q$, the layer truncation ratio $h/L$, and the average layer-wise budget $m$, i.e., $\text{RR} = p \times q \times \frac{h}{L} \times m$.
For example, $p=0.75$, $q=0.67$, $h=\frac{3}{4}L$, and $m=0.75$ together yield $\text{RR}\approx 28\%$.
For baseline methods, we tune their respective hyperparameters to match the same global RR.
The determination of $h$ and $\mu$ is further analyzed in Appendix~\ref{truncated_layer_infer_performance} and Appendix~\ref{impact_fusion_strength}. 
In addition, the scout VLM is model-agnostic, as discussed in Appendix~\ref{impactvlm}.


\begin{table*}[!t]
  \centering
  \caption{
    {Results on MVBench of different KV cache compression methods with $(p=0.75, q=0.67)$.}    
    The first row lists abbreviations of the subset names, from \textit{AS} to \textit{CI} (e.g., Action Sequence (AS)).
    Bold and underlined numbers indicate the best and second-best results, respectively.
  }
  \resizebox{\linewidth}{!}{
  \begin{tabular}{lccccccccccccccccccccccccc}
    \toprule  
    Method & RR & AS & AP & AA & FA & UA & OE & OI & OS & MD & AL & ST & AC & MC & MA & SC & FP & CO & EN & ER & CI & Avg \\
    \midrule
    \multicolumn{23}{c}{ Qwen2.5-VL-3B-Instruct} \\
    \midrule
    Full Cache & 100\% & 69.0 & 66.8 & 72.0 & 26.4 & 72.4 & 85.0 & 69.7 & 31.3 & 50.5 & 37.5 & 76.0 & 48.5 & 62.5 & 89.0 & 45.5 & 27.0 & 47.1 & 36.5 & 26.8 & 58.0 & 56.6 \\
    Q-Frame & 50\% & 49.5 & 63.3 & \underline{68.5} & 23.4 & \underline{64.3} & \textbf{80.5} & 59.6 & 31.3 & \textbf{46.0} & 37.5 & \textbf{77.0} & 36.4 & 56.5 & 78.0 & 40.9 & \underline{24.5} & \textbf{52.9} & \textbf{35.0} & \textbf{37.0} & 61.5 & 51.2 \\
    PyramidKV & 50\% & \textbf{65.5} & \textbf{66.8} & 54.0 & 24.9 & 46.4 & \underline{77.5} & 55.6 & \textbf{43.8} & 39.0 & \underline{38.0} & 51.0 & \underline{42.4} & \underline{62.5} & \textbf{87.0} & 45.5 & 20.5 & \textbf{52.9} & 32.0 & \underline{28.8} & 53.0 & 49.4 \\
    FastV & 50\% & 60.5 & 53.8 & 61.5 & \textbf{29.4} & 48.0 & \textbf{80.5} & \underline{64.7} & 18.8 & \underline{41.0} & \textbf{38.5} & 50.5 & 36.4 & \textbf{63.0} & 81.5 & 45.5 & 20.5 & 47.1 & 30.0 & 22.5 & \textbf{62.5} & 47.8 \\
    PDrop & 50\% & 54.5 & 58.3 & \textbf{72.5} & 24.4 & \textbf{65.5} & \underline{77.5} & 63.7 & 30.0 & 44.5 & 36.0 & 70.5 & \textbf{45.5} & \textbf{63.0} & \underline{86.5} & \underline{50.0} & \textbf{29.0} & \underline{50.0} & \underline{34.5} & 25.0 & \underline{53.5} & \underline{51.7} \\
    \rowcolor{blue!10}
    VisCache & 50\% & \underline{62.5} & \underline{63.8} & 63.0 & \underline{27.9} & 62.6 & 71.0 & \textbf{68.7} & \underline{37.5} & 40.5 & 37.5 & \underline{76.5} & \underline{42.4} & 61.5 & 83.5 & \textbf{68.8} & \underline{24.5} & 47.1 & 32.7 & 27.3 & \underline{53.5} & \textbf{52.6}\\
    \midrule  
    \multicolumn{23}{c}{Qwen2.5-VL-32B-Instruct} \\
    \midrule
    Full Cache & 100\% & 68.0 & 66.3 & 76.0 & 39.5 & 70.2 & 84.0 & 69.5 & 68.8 & 44.0 & 39.0 & 85.0 & 39.4 & 57.0 & 79.0 & 63.6 & 35.0 & 47.1 & 38.0 & 45.5 & 46.5 & 58.1 \\
    Q-Frame & 50\% & \underline{49.0} & \underline{64.3} & \underline{72.5} & 31.0 & 58.8 & 76.0 & \textbf{63.6} & \textbf{75.0} & 44.5 & \textbf{37.5} & 81.0 & 33.3 & 45.0 & \underline{74.5} & 45.5 & 23.0 & \underline{52.9} & 20.0 & 44.5 & 45.5 & 51.9 \\
    PyramidKV & 50\% & 43.2 & 56.7 & \textbf{80.5} & 33.5 & 66.3 & 74.0 & 59.0 & 46.7 & 43.5 & 19.9 & \textbf{85.5} & \textbf{39.4} & \textbf{51.5} & 66.0 & 54.6 & 25.5 & \textbf{58.8} & 25.0 & \underline{41.2} & \textbf{49.0} & 51.0 \\
    FastV & 50\% & 34.7 & 54.2 & 71.5 & 37.1 & 65.1 & 70.5 & 53.2 & 56.3 & 39.5 & \underline{33.0} & 72.5 & 30.3 & 50.0 & \textbf{78.0} & 45.5 & \underline{34.0} & \underline{52.9} & \textbf{37.0} & 36.0 & 46.0 & 49.9 \\
    PDrop & 50\% & \underline{49.0} & 52.3 & \textbf{80.5} & \underline{38.6} & \textbf{77.7} & \underline{81.0} & 57.8 & \underline{62.5} & \textbf{51.5} & \underline{33.0} & \underline{84.5} & 30.3 & 47.0 & 72.0 & \underline{63.6} & \textbf{41.5} & 35.3 & \underline{36.5} & \textbf{43.0} & 46.0 & \underline{54.9} \\
    \rowcolor{blue!10}
    VisCache & 50\% & \textbf{51.9} & \textbf{64.8} & 72.0 & \textbf{41.6} & \underline{74.4} & \textbf{84.5} & \underline{61.0} & 56.3 & \underline{51.0} & 30.5 & \textbf{85.5} & \underline{36.4} & \underline{51.0} & 69.0 & \textbf{68.8} & 32.0 & \underline{52.9} & 36.4 & 39.6 & \underline{48.0} & \textbf{55.4} \\
    \bottomrule
  \end{tabular}
  }
  \label{MVBench1}
\end{table*}




\subsection{Main Results}
Table~\ref{tab:main_result1} compares VisCache with competitive KV cache compression baselines under controlled retention ratios. VisCache consistently achieves the lowest FLOPs across all settings. At 40\% RR, it reduces FLOPs to 9\% of the full cache on the 3B model and 15\% on the 32B model, outperforming the next best method PyramidKV. This advantage stems from the dual-stage design: the scout filters redundant frames before inference, while PruneKV further compresses the KV cache along both token and layer dimensions. At more aggressive RRs (28\% and 19\%), FLOPs drop to 7\% and 6\% on the 3B model, demonstrating the scalability of our framework.

Despite aggressive compression, VisCache maintains strong accuracy. On the 3B model, VisCache at 28\% RR achieves the highest average accuracy (45.64\%), surpassing the full cache (44.16\%) and all baselines. On the 32B model, it achieves the best average accuracy (56.86\%) while discarding 72\% of the KV cache. VisCache consistently ranks first or second on NExTQA, ActQA, and EgoSchema. On VS benchmarks, performance is comparable across methods, with only a slight trade-off on ActCap under extreme compression—acceptable given the substantial efficiency gains. Notably, while PyramidKV performs well at 40\% RR, it degrades substantially under more aggressive compression (see Appendix~\ref{app:pyramid_28}), confirming that parabolic allocation better preserves essential information when the budget is tight.

Varying RR from 40\% to 19\% reveals a clear accuracy--efficiency spectrum: 40\% prioritizes accuracy, 28\% offers the best overall trade-off, and 19\% still retains strong performance, outperforming most baselines at higher RRs. These results highlight VisCache's robustness under extreme compression, where uniform or arithmetic budgets often collapse. 
Detailed comparisons at additional RRs are provided in Appendix~\ref{app:extra_rr}. VisCache scales effectively from 3B to 32B: efficiency gains are consistent, and accuracy advantages grow with model size. On the 32B model, VisCache at 28\% RR outperforms all baselines at 40\% RR.

Table~\ref{MVBench1} reports the performance on MVBench, a comprehensive multi-task benchmark spanning 20 reasoning tasks.
VisCache achieves the highest average accuracy on both model scales (52.6\% on 3B, 55.4\% on 32B), surpassing all baselines at the same retention ratio.
On the 32B model, VisCache outperforms the strongest baseline PDrop by 0.5 points in average accuracy, with particularly strong gains on tasks requiring temporal reasoning (AS: 51.9\%, UA: 74.4\%) and fine-grained perception (AA: 41.6\%).
Notably, VisCache and PDrop exhibit complementary strengths across sub-tasks: VisCache leads on 11 out of 20 tasks on the 32B model, while PDrop leads on the remaining tasks, suggesting that parabolic allocation and asymmetric fusion are especially beneficial for tasks that depend on structured visual information retained across layers.
The overall margin over the full cache is small (within 3 points on 32B), confirming that VisCache preserves broad reasoning capabilities under aggressive compression.
The compatibility evaluation of VisCache is provided in Appendix~\ref{compatiable} and Appendix~\ref{Combine with quantization}. 
In addition, representative inference examples are presented in Appendix~\ref{Examples}.

\subsection{Analysis and Ablation Studies}
\begin{table}[!t]
\centering
\caption{
  Real-system inference efficiency on ActCap.
  \textit{Mem.}: GPU memory in GB (Total = peak usage, KV = raw KV cache size).
  \textit{TPS}: tokens per second (higher is better).
}
\label{tab:memory_compare}
\small
\resizebox{\linewidth}{!}{
\begin{tabular}{l|c|c|c|c}
\toprule
Method & RR & Total Mem. & KV Cache & TPS \\
\midrule
Full Cache & 100\% & 5.57 & 0.06 & 10.4 \\
PyramidKV  & 40\%  & 4.98 & 0.02 & 9.7  \\
FastV      & 40\%  & 4.98 & 0.02 & 10.6 \\
PDrop      & 40\%  & 4.99 & 0.02 & 15.0 \\
\midrule
VisCache   & 40\%  & \textbf{4.10} & 0.03 & 15.5 \\
VisCache   & 28\%  & \textbf{4.10} & 0.02 & \textbf{18.6} \\
VisCache   & 19\%  & \textbf{3.68} & 0.02 & 15.7 \\
\bottomrule
\end{tabular}
}
\end{table}
\paragraph{Memory Usage.}
We measure the total GPU memory and the KV cache GPU memory for different acceleration methods on ActCap based on Qwen2.5-VL-3B-Instruct backbone. As Table~\ref{tab:memory_compare} shown, VisCache reduces the total GPU memory and KV cache GPU memory to 3.68 GB and 0.02 GB while maintaining competitive throughput and nearly identical inference performance compared to the baseline methods.
These results demonstrate that VisCache provides an efficient trade-off between memory consumption and inference speed, enabling deployment of VLLMs on resource-constrained edge devices without incurring significant degradation in output quality.
\begin{figure}[!t]
    \centering
    \includegraphics[width=\linewidth]{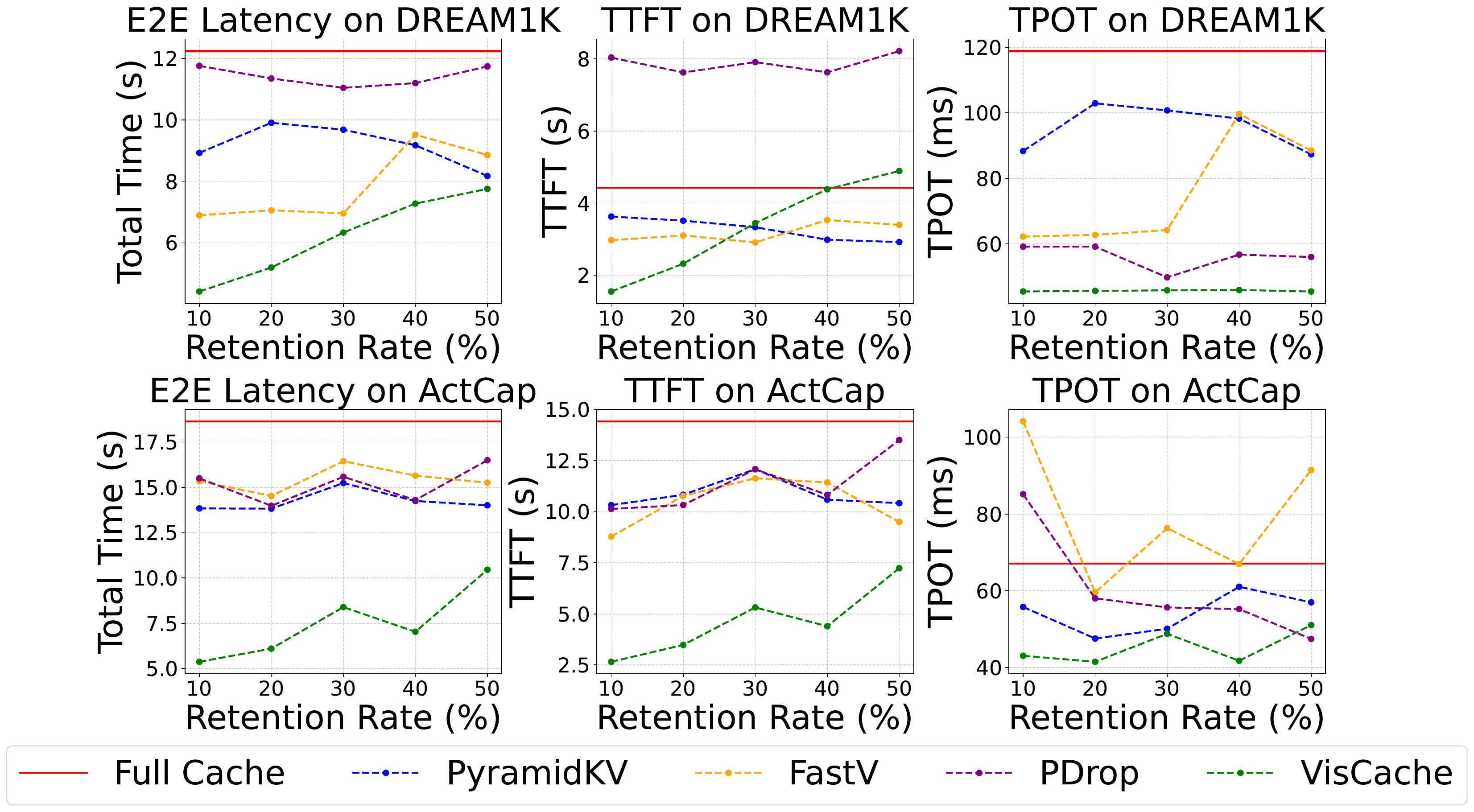}
    \caption{
    Comparison of average inference time of different baseline algorithms on Qwen2.5-VL-3B-Instruct at different KV cache compression rates for the VS task.
    The left, middle, and right columns report the average end-to-end (E2E) latency, time to first token (TTFT), and time per output token (TPOT), respectively.
    }   
    \label{Inference Time}
\end{figure}
\begin{figure}[!t]
    \centering
    \includegraphics[width=0.8\linewidth]{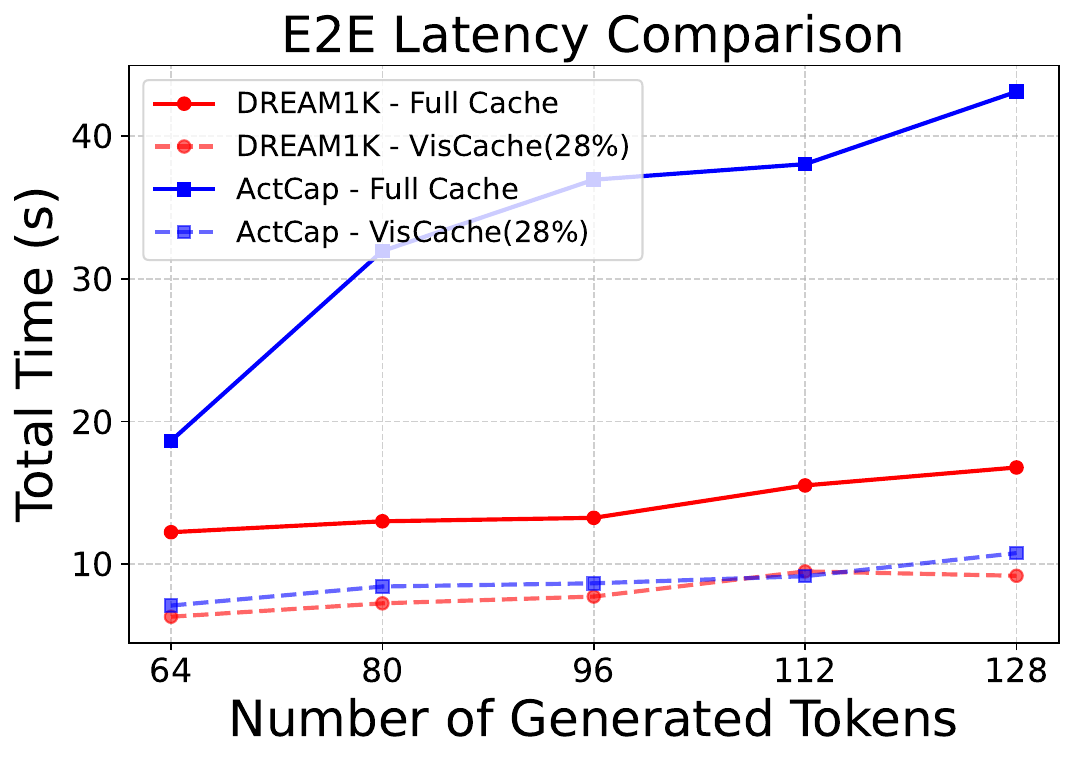}
    \caption{
    {Comparison of average inference time between full KV cache and VisCache under varying numbers of output tokens.}
    }    
    \label{Decode Time}
\end{figure}
\paragraph{Inference Time.}

We record the computation time of VisCache and baseline algorithms under different visual KV cache compression rates and plot the results in Figure~\ref{Inference Time}.
It can be clearly observed that the average End-to-End (E2E) inference time of VisCache-which includes the keyframe selection stage as well as the subsequent VLLM inference-remains consistently lower than that of the full cache and baseline methods across all retention rates, further validating the overall efficiency of VisCache.
By retaining just 28$\%$ of the KV cache, VisCache already achieves impressive E2E speedups of up to 1.93 $\times$ over the full cache setting.
Reducing the retention further to 19$\%$ amplifies the effect, pushing the acceleration to an astonishing 2.35 $\times$, all while achieving performance close to the baseline.
Specific inference time measurements are explained in Appendix~\ref{specific_time_analysis}.

VisCache significantly reduces Time Per Output Token (TPOT), yielding substantial time savings compared to full cache and baseline methods, while its effect on Time To First Token (TTFT) is minimal.
This indicates that although the first token is not noticeably accelerated, subsequent decoding is greatly sped up, more than compensating for the initial overhead and reducing overall E2E latency.
To better illustrate this phenomenon, We increase the number of output tokens in the model from 64 to 128, and the experimental results are shown in Figure~\ref{Decode Time}. 
The more tokens output, the more time VisCache saves compared to full cache, further confirming our conclusion.

\begin{table}[!t]
    \centering
    \caption{
        Ablation study on budget allocation strategy and value cache fusion.
        We keep $\text{RR}=28\%$ of the visual KV cache on Qwen2.5-VL-32B-Instruct across all variants.
        \textit{Fixed}: uniform budget per layer.
        \textit{Arithmetic}: arithmetic progression (PyramidKV~\cite{cai2024pyramidkv}).
        \textit{Geometric}: geometric progression across four layer blocks (PDrop~\cite{xing2024pyramiddrop}).
        \textit{Parabola}: our proposed parabolic schedule (Equation~\ref{eq:parabolic}).
        \textit{V Cache Fusion}: asymmetric value fusion.
    }
    \label{tab:ablation}
    \resizebox{\linewidth}{!}{
        \begin{tabular}{l|c|cc|cccc}
        \toprule
        Budget & V Cache & ActCap & DREAM1K & NExTQA & ActQA & EgoSchema & Avg.  \\
        Strategy & Fusion & ROUGE-L & ROUGE-L & Acc & Acc & Acc & Acc \\
        \midrule
        \multicolumn{8}{c}{\textit{Full cache (upper bound)}} \\
        Full Cache & -- & 2.82 & 7.87 & 60.71 & 46.47 & 65.20 & 57.46 \\
        \midrule
        \multicolumn{8}{c}{\textit{Without V Cache Fusion}} \\
        Fixed      & $\times$ & 2.43 & 7.24 & 59.00 & 43.44 & 63.60 & 55.35 \\
        Arithmetic  & $\times$ & 2.67 & 7.17 & 58.07 & 42.31 & 57.00 & 52.46 \\
        Geometric   & $\times$ & 2.69 & 7.25 & 55.56 & 42.68 & 57.60 & 51.95 \\
        {Parabola} & $\times$ & 2.70 & 7.05 & {61.61} & {42.64} & {65.00} & {56.42} \\
        \midrule
        \multicolumn{8}{c}{\textit{With V Cache Fusion}} \\
        Fixed      & \checkmark & 2.69 & 7.36 & 61.58 & 42.42 & 64.80 & 56.27 \\
        Arithmetic  & \checkmark & 2.72 & 7.23 & 61.32 & 42.33 & 65.00 & 56.22 \\
        Geometric   & \checkmark & 2.69 & 7.27 & 56.07 & 40.49 & 56.00 & 50.85 \\
        {Parabola} & \checkmark & 2.72 & 7.29 & {{62.08}} & {{42.69}} & {{65.80}} & {{56.86}} \\
        \bottomrule
        \end{tabular}
    }
\end{table}

\paragraph{Impact of Layer-wise Budget Allocation and Value Fusion.}
We first ablate the two core design choices in PruneKV: the layer-wise budget allocation strategy and the asymmetric value fusion mechanism.
Table~\ref{tab:ablation} reports results under a fixed 28\% KV cache retention ratio.
Among the four budget strategies, Parabola consistently achieves the best performance across all three VQA benchmarks, with its average accuracy approaching that of the full cache upper bound.
Arithmetic and Fixed strategies also perform competitively, while Geometric allocation degrades substantially, confirming that overly aggressive compression in early layers impairs fine-grained visual understanding.
To assess the contribution of value fusion, we compare each budget strategy with and without fusion.
For Parabola, Fixed, and Arithmetic allocations, enabling fusion consistently improves accuracy, demonstrating that asymmetric value fusion effectively preserves contextual information that would otherwise be lost.
The exception is Geometric allocation, where fusion slightly degrades performance---likely because excessive pruning leaves too few retained tokens for meaningful redistribution, further underscoring the importance of pairing fusion with a well-designed budget strategy.
Notably, Parabola with fusion achieves the best individual scores across all VQA benchmarks, validating that the two components are synergistic: parabolic allocation retains tokens where they matter most, while value fusion salvages information from pruned tokens.

\subsection{Ablation on Two Stages of VisCache.}
\label{mutlistage_Viscache}
\begin{table}[!t]
    \centering
    \caption{{Comparison of different KV cache compression methods on various VQA and VS datasets.} 
    Bold and underlined numbers indicate the best and second-best results, respectively.
    }
    \resizebox{\linewidth}{!}{
    \begin{tabular}{l|c|c|c|cc}
    \toprule  Methods & RR & Scout & PruneKV & DREAM1K & EgoSchema \\
    &    &       &        & ROUGE-L & Acc \\
    \midrule    
    PDrop & 40\% & - & - & 8.63 & 50.00 \\
    VisCache & 40\% & $\surd$ & $\times$ & 8.48 & 46.20 \\
    VisCache & 40\% & $\times$ & $\surd$ & 9.65 & 48.00 \\
    VisCache & 40\% & $\surd$ & $\surd$ & 9.79 & 52.00 \\
    \midrule
    PDrop & 28\% & - & - & 8.62 & 49.60 \\ 
    VisCache & 28\% & $\surd$ & $\times$ & 8.13 & 46.00 \\ 
    VisCache & 28\% & $\times$ & $\surd$ & 8.62 & 48.80 \\ 
    VisCache & 28\% & $\surd$ & $\surd$ & 8.70 & 55.00 \\
    \midrule
    PDrop & 19\% & - & - & 7.69 & 49.00 \\
    VisCache & 19\% & $\surd$ & $\times$ & 7.84 & 48.20 \\
    VisCache & 19\% & $\times$ & $\surd$ & 8.23 & 48.60 \\ 
    VisCache & 19\% & $\surd$ & $\surd$ & 8.26 & 54.60 \\
    \bottomrule
    \end{tabular}
    }
    \label{app:ablation_results}
\end{table}
Table~\ref{app:ablation_results} presents the ablation study of the two-stage design in VisCache, including Scout-based temporal redundancy filtering and PruneKV-based visual KV cache compression.
We first observe that applying only Scout-based filtering leads to noticeable performance degradation in some settings, especially on EgoSchema, indicating that solely removing temporally redundant frames may discard informative visual context required for reasoning.
In contrast, using only PruneKV generally achieves more stable performance and consistently outperforms the baseline PDrop under aggressive compression ratios, demonstrating the effectiveness of fine-grained visual KV cache pruning.
\begin{table}[!t]
    \centering
    \caption{Ablation of shared global and layer-specific visual-token ranking at matched retention ratios on Qwen2.5-VL-3B-Instruct. 
    Bold numbers indicate the best result under each retention ratio.}
    \label{tab:shared_ranking}
    \resizebox{\linewidth}{!}{%
    \begin{tabular}{l|c|cc|cccc}
        \toprule
        Ranking strategy & RR
        & ActCap & DREAM1K
        & NExTQA & ActQA & EgoSchema & Avg.\ Acc. \\
        & & ROUGE-L & ROUGE-L & Acc. & Acc. & Acc. & \\
        \midrule
        Shared global ranking
        & 28\%
        & \textbf{2.45} & 8.70
        & \textbf{41.25} & \textbf{40.66}
        & \textbf{55.00} & \textbf{45.64} \\
        
        Layer-specific ranking
        & 28\%
        & 2.35 & \textbf{8.73}
        & 41.20 & 39.82
        & 51.75 & 44.26 \\
        \midrule
        
        Shared global ranking
        & 19\%
        & \textbf{2.39} & \textbf{8.26}
        & \textbf{41.44} & \textbf{38.51}
        & \textbf{54.60} & \textbf{44.85} \\
        
        Layer-specific ranking
        & 19\%
        & 2.29 & 7.95
        & 41.24 & 37.41
        & 53.20 & 43.95 \\
        \bottomrule
    \end{tabular}%
    }
\end{table}
More importantly, combining both Scout and PruneKV consistently achieves the best overall performance across different RRs.
This suggests that the two stages are highly complementary: Scout reduces coarse-grained temporal redundancy at the frame level, while PruneKV further refines the retained information through token-level KV cache compression.
The advantage becomes more significant at lower RRs, where the complete VisCache framework maintains strong reasoning performance even under highly constrained memory budgets.

\subsection{Ablation on Shared vs.\ Layer-Specific Visual Token Ranking}
\label{shared_ranking}
We further examine whether visual token importance should be computed globally across all layers or independently for each layer.
In the shared global ranking, the importance of visual token \(v\) is obtained by aggregating its attention scores across all transformer layers:
\begin{equation}
    s_v = \sum_{l=1}^{L} \sum_{i} A_{i,v}^{(l)},
\end{equation}
where \(A_{i,v}^{(l)}\) denotes the attention weight received by visual token \(v\) at layer \(l\). All layers then select tokens from the same global ordering according to their retention budgets, i.e., layer \(l\) keeps the top-\(k_l\) tokens from this shared ranking. In the layer-specific variant, each layer independently computes
\begin{equation}
    s_v^{(l)} = \sum_{i} A_{i,v}^{(l)}
\end{equation}
and selects its own top-\(k_l\) visual tokens.
Table~\ref{tab:shared_ranking} compares the two strategies at matched retention ratios on Qwen2.5-VL-3B-Instruct. Shared global ranking improves the average VQA accuracy from 44.26 to 45.64 at 28\% retention ratio and from 43.95 to 44.85 at 19\% retention ratio. It also achieves the best result on most individual benchmarks. This design intentionally decouples {which} tokens are important from {how many} tokens each layer retains.
Aggregating attention across all layers provides a more stable and comprehensive importance estimate, while the parabolic budget still assigns different retention cardinalities to each layer.
In contrast, layer-specific ranking may select substantially different token identities across layers, which can reduce the consistency of visual information preserved throughout the backbone.

\section{Related Work}

Post-aligned LLMs and VLLMs usually come with redundant responses~\citep{li-etal-2025-self,li-etal-2025-aplot,ICLR2026_d2f6f1df}, leading to additional unsafe behaviors~\citep{wang-etal-2026-safeguarding,du-etal-2025-atoxia,li-etal-2026-march} and inference cost. Existing efficient VLLM inference methods improve efficiency by localizing query-aware video clip, reducing visual token redundancy or compressing KV caches. 
SeViLA~\cite{yu2023self} performs query-aware keyframe localization and QA via a self-chained design, while LongVU~\cite{shen2024longvu} achieves efficient long-video understanding via adaptive spatiotemporal token compression.
One line of work focuses on token pruning and merging based on attention or semantic importance. Methods such as FastV~\cite{chen2024image}, VisionZip~\cite{yang2025visionzip}, and SparseVLM~\cite{zhang2024sparsevlm} select or merge tokens using attention or cross-modal relevance, while approaches like FitPrune~\cite{ye2025fit}, PruMerge~\cite{shang2025llava}, and DivPrune~\cite{alvar2025divprune} further explore optimization and diversity-aware selection strategies. Another line of work reduces inference cost via structured KV cache compression or layer-wise budget allocation. PyramidKV~\cite{cai2024pyramidkv}, PyramidInfer~\cite{yang2024pyramidinfer}, and PDrop~\cite{xing2024pyramiddrop} allocate KV budgets across layers using predefined schedules such as arithmetic or geometric decay. Despite their effectiveness, these methods typically treat tokens or KV caches in a coarse-grained manner, ignoring the heterogeneous importance of visual information across layers and the asymmetric roles of keys and values, which can limit compression performance under aggressive budgets.

\section{Conclusion}
We present {VisCache}, a training-free framework for efficient visual KV cache compression in long-video VLLMs via prompt-aware filtering and layer-aware {PruneKV}, preserving hierarchical context while improving long-context video efficiency.

\section*{Limitations}

We identify two primary limitations of this work.
First, the scout-based temporal filtering stage relies on a lightweight VLM such as CLIP, whose visual encoder may not perfectly align with the main LLM backbone.
In scenarios where the scout and the main model exhibit substantially different visual representations, the selected keyframes may deviate from what the downstream model would consider optimal.
Second, the current implementation of PruneKV requires storing attention scores from all layers during the prefilling stage, which introduces some memory overhead.
We leave the exploration of scout‑backbone co‑adaptation and memory‑efficient score computation to future work.

\section*{Statement of Impacts}

This work aims to improve the inference efficiency of vision large language models, thereby lowering the computational barrier for deploying advanced video understanding capabilities in practice.
By substantially reducing memory consumption and latency, VisCache can contribute to broader accessibility of VLLM technologies, particularly in academic and resource-limited settings.

We acknowledge that efficiency improvements in VLLMs may also facilitate large-scale video analysis applications, including automated surveillance and mass media monitoring.
We encourage practitioners deploying VisCache in such contexts to adhere to established ethical guidelines regarding privacy, consent, and algorithmic fairness.
Our method reduces visual token redundancy based on learned attention patterns from pretrained models; as such, any biases present in the underlying foundation models may propagate through the compression process.
We recommend that users of VisCache conduct fairness and bias assessments tailored to their specific application domains.

\section*{Acknowledgments}
This work was supported in part by the National Natural Science Foundation of China (Grant No. 62522118, 62371313), in part by the Shenzhen Science and Technology Program (Grant No. JCYJ20241202124934046), in part by the Guangdong Young Talent Research Project (Grant No. 2023TQ07A708), in part by Shenzhen Loop Area Institute (Contract No. SLAI2026020007).

\bibliography{custom}

\newpage
\appendix
\section{LLM Usage Statement}
We employ a large language model (LLM) as a general-purpose writing assistant to improve the clarity and fluency of the text. Its role is limited to refining linguistic expression and enhancing overall readability and coherence.

\section{Algorithm}
\label{app:algorithm}

\begin{algorithm}[!h]
\caption{PruneKV: Layer-Aware Visual KV Compression}
\label{alg:prunekv}
\begin{algorithmic}[1]
\Require Visual KV cache $\{(\mathbf{K}^l, \mathbf{V}^l)\}_{l=1}^{L}$, attention scores $\{\mathbf{A}^l\}_{l=1}^{L}$, truncation layer $h$, retention ratio $m$, fusion strength $\mu$, temperature $\tau$
\Ensure Compressed visual KV cache $\{(\mathbf{K}_{\text{new}}^l, \mathbf{V}_{\text{new}}^l)\}_{l=1}^{h}$
\State \Comment{Step 1: Token-level importance scoring}
\State $\bar{\mathbf{A}} \leftarrow \frac{1}{L} \sum_{l=1}^{L} \mathbf{A}^l$  \Comment{Average attention across layers}
\State $s_v \leftarrow \sum_{i=1}^{N} \bar{A}_{i,v}, \quad \forall v$ \Comment{Aggregate attention received by each visual token}
\State Retain KV entries of the $q$ percent visual tokens by $s_v$, discard the rest \Comment{$r$ determined by $m$ and $h$}
\State \Comment{Step 2: Parabolic budget allocation}
\State Compute $\{b_l\}_{l=1}^{h}$ via Equation~\eqref{eq:parabolic}
\State \Comment{Step 3: Layer-wise asymmetric compression}
\For{$l \leftarrow 1$ \textbf{to} $h$}
    \State Partition visual tokens at layer $l$ into keep set $\mathcal{C}_k$ (top-$b_l$ by $s_v$) and drop set $\mathcal{C}_d$
    \State $\mathbf{K}_{\text{new}}^l \leftarrow$ key vectors of $\mathcal{C}_k$ \Comment{Prune keys in $\mathcal{C}_d$}
    \State $\mathbf{V}_k \leftarrow$ value vectors of $\mathcal{C}_k$, $\mathbf{V}_d \leftarrow$ value vectors of $\mathcal{C}_d$
    \State Compute redistribution matrix $\mathbf{\Phi}$ via Equation~\eqref{eq:redistribution} \Comment{Similarity-based affinity}
    \State $\mathbf{V}_{\text{new}}^l \leftarrow \mu \mathbf{V}_k + (1-\mu)\, \mathbf{\Phi} \mathbf{V}_d$ \Comment{Fuse dropped values into kept values, Equation~\eqref{eq:weight_fusion}}
\EndFor
\State \Return $\{(\mathbf{K}_{\text{new}}^l, \mathbf{V}_{\text{new}}^l)\}_{l=1}^{h}$
\end{algorithmic}
\end{algorithm}

\section{The Effect of Selecting Keyframes by Small VLM}
\label{The Effect of Selecting Keyframes by Small VLM}
\begin{figure}[ht]
    \centering
    \includegraphics[width=0.4\textwidth]{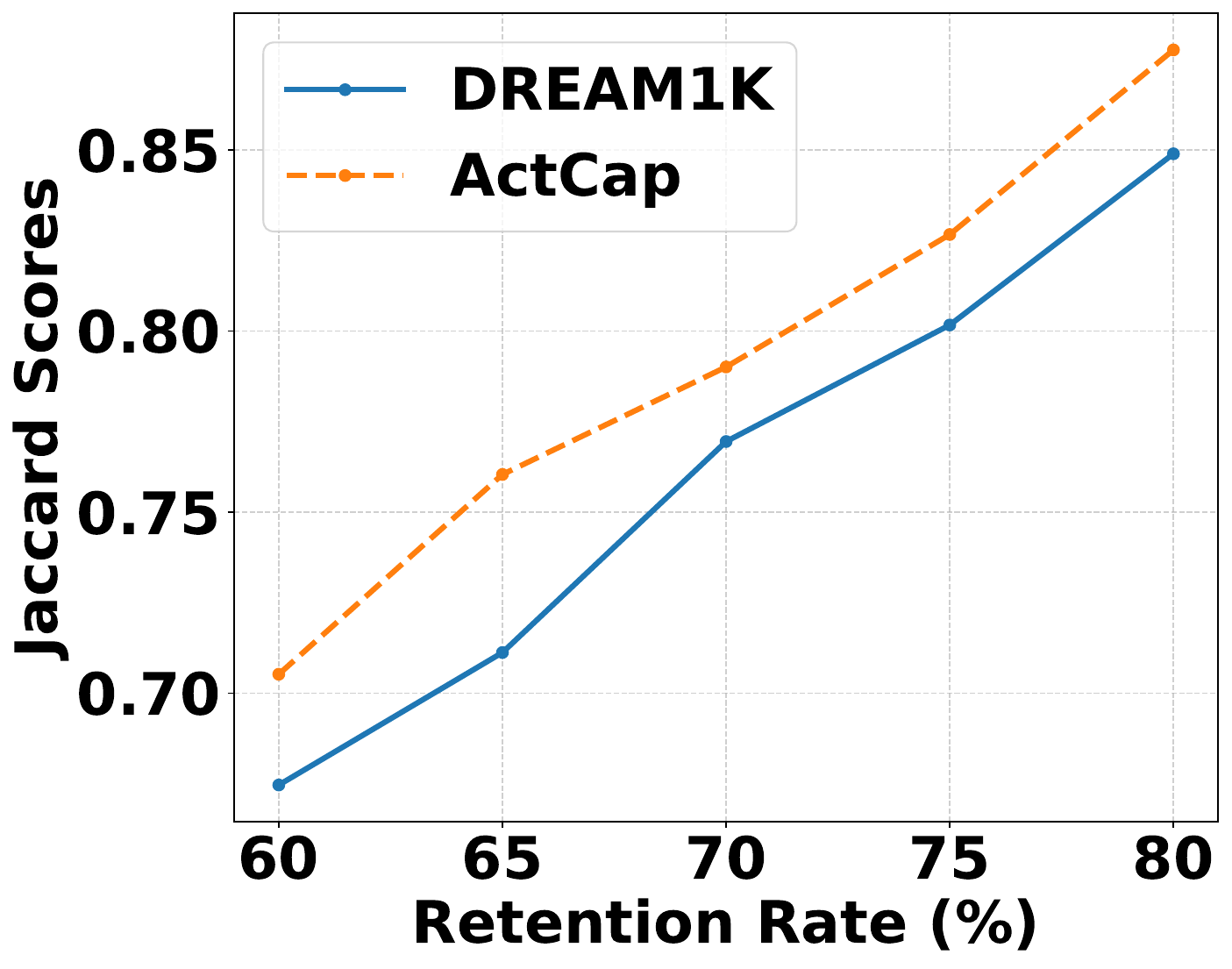}
    \caption{
    {Jaccard similarity between key visual tokens selected by the small VLM with MMR and those selected from the sum of attention scores across Qwen2.5-VL-3B-Instruct layers, under different visual token compression rates.}
    This figure highlights the consistency between the two token sets at varying compression levels.
    }    
    \label{Jaccard_Scores_Compression}      
\end{figure}

To examine the reliability of using a small VLM for keyframe selection, we conduct the following experiment.
During the prefilling stage, we identify a set of key visual tokens by aggregating attention scores across all layers of the VLLM.
We then compare this token set with the visual tokens contained in keyframes selected by a small VLM guided by MMR-based selection.
Specifically, on two generation-oriented benchmarks, we compute the Jaccard similarity between these two token sets while ensuring they contain the same number of tokens.
By varying the ratio between the selected tokens and the total visual tokens in the original video, we obtain the results illustrated in the Figure~\ref{Jaccard_Scores_Compression}.

The results show that, across different compression ratios, the visual token sets selected by the small VLM combined with the MMR-based algorithm consistently exhibit high Jaccard similarity with the tokens receiving higher attention scores in the VLLM.
This indicates that the keyframes identified by the small VLM largely correspond to the visually salient tokens emphasized by the VLLM itself.
By filtering these keyframes in advance, the number of visual tokens entering the prefilling stage is substantially reduced, thereby lowering the computational cost of both prefilling and decoding and ultimately improving the inference efficiency of the VLLM.

\section{How To Determine the Truncation Layer of VLLM}
\label{truncated_layer_infer_performance}
To determine an appropriate truncation layer for VLLM, we jointly consider the trade-off between inference accuracy and inference efficiency. 
Specifically, we conduct an ablation study on Qwen2.5-VL-32B-Instruct, which contains 64 transformer layers in total. 
For different truncation points, we evaluate the inference performance on EgoSchema while also measuring the corresponding inference speed.

\begin{figure}[!t]
    \centering
    \includegraphics[width=0.9\linewidth]{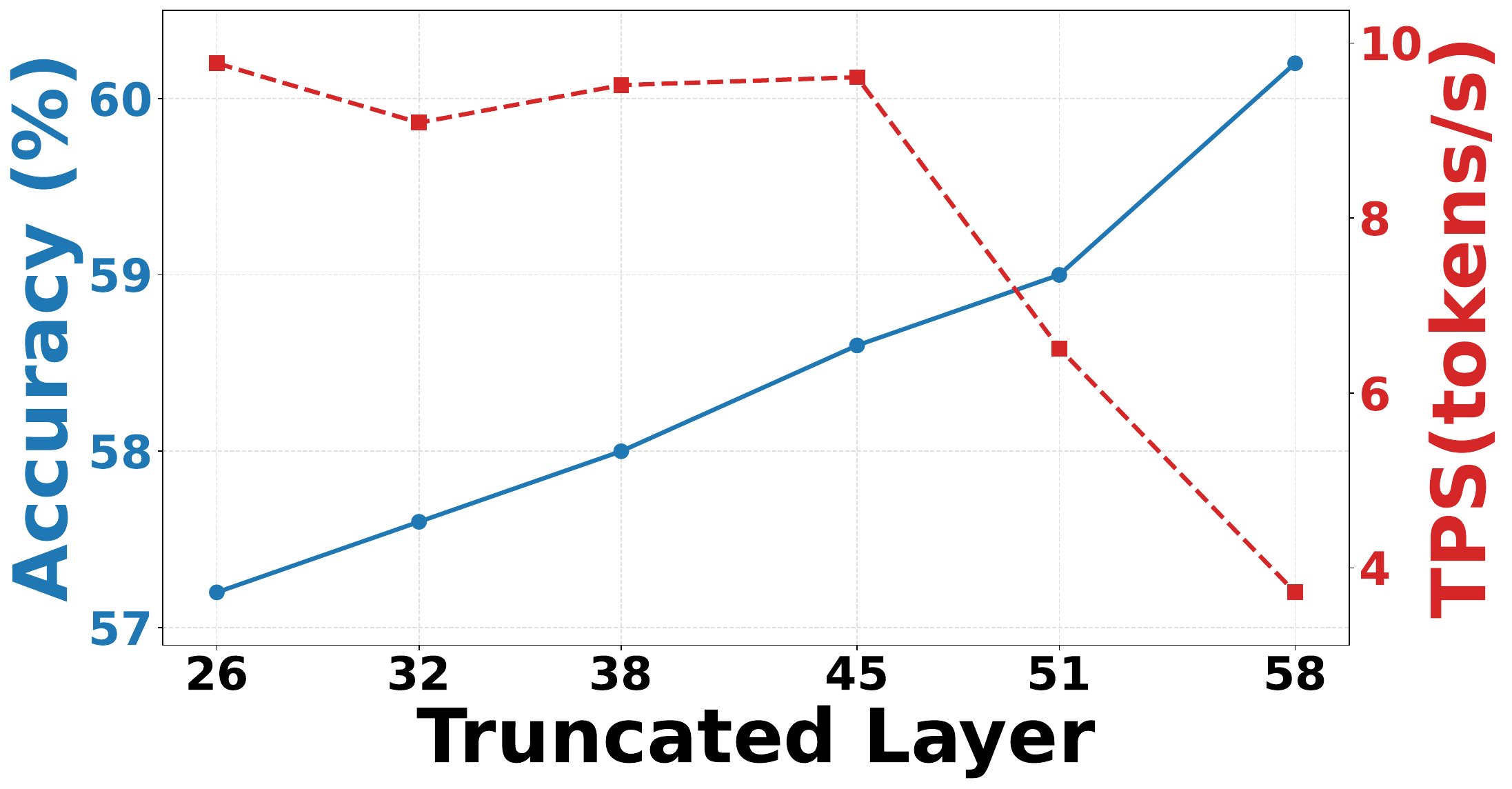}
    \caption{Trade-off between inference accuracy and inference speed under different truncation layers of VLLM.}
    \label{truncated_layer_acc}
\end{figure}

As illustrated in Figure~\ref{truncated_layer_acc}, increasing the truncation layer generally improves inference accuracy because more high-level semantic information can be preserved during decoding. 
However, this also introduces higher computational overhead, resulting in slower inference speed. 
In contrast, truncating at earlier layers significantly accelerates inference but causes noticeable performance degradation due to insufficient semantic reasoning capability.

From the experimental results, we observe that the balance between inference
accuracy and speed is reached when the truncation layer is set to 48, which
corresponds to $3/4$ of the total model depth (64 layers).
At this point, the VLLM maintains strong reasoning capability while avoiding the
substantial latency increase introduced by deeper layers.
Therefore, we adopt the $3/4$ truncation depth as the default configuration in VisCache.

\section{Impact of Fusion Strength}
\label{impact_fusion_strength}
\begin{figure}[!h]
    \centering
    \includegraphics[width=0.9\linewidth]{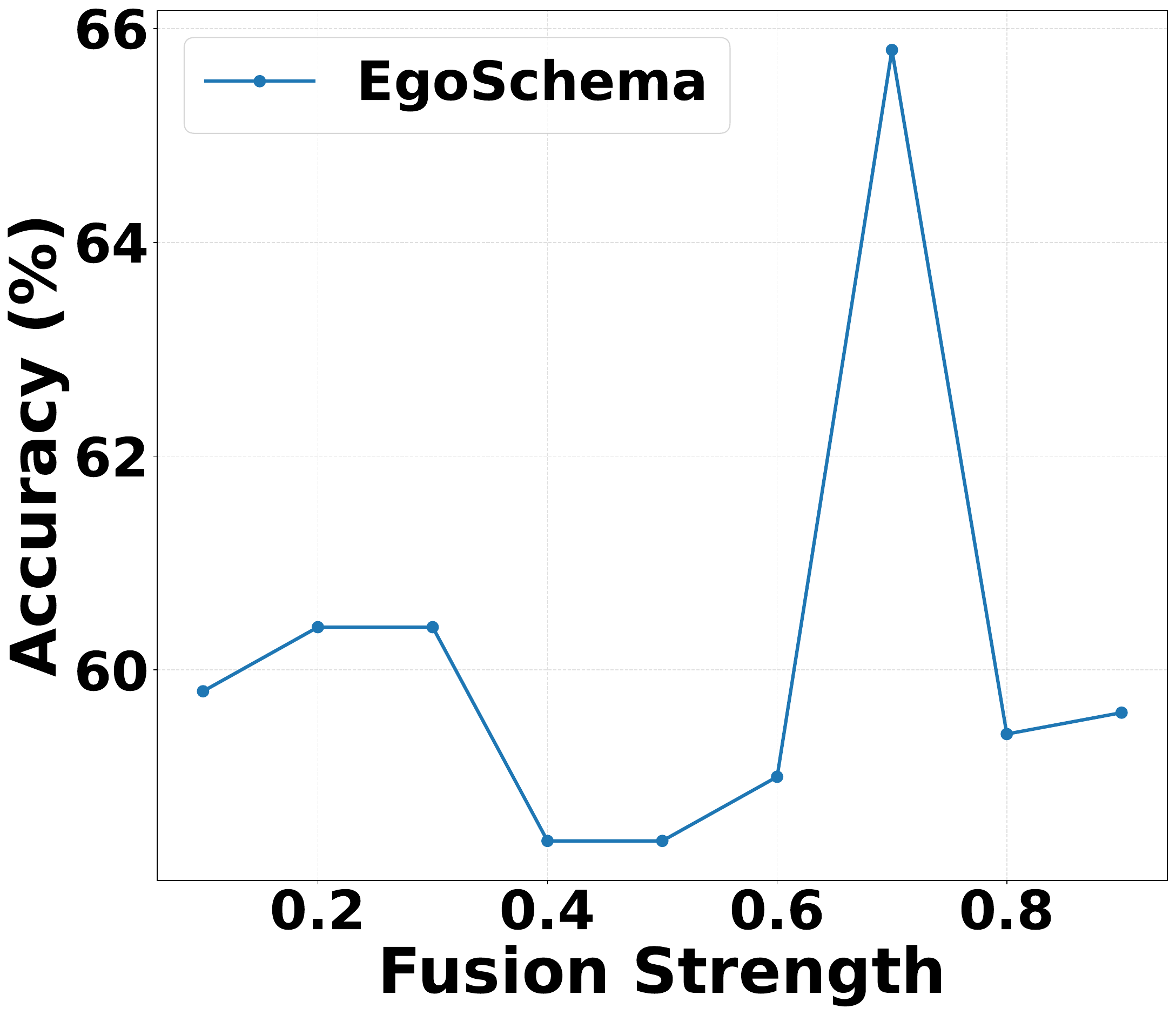}
    \caption{{The relationship between fusion strength $\mu$ and accuracy on EgoSchema with Qwen2.5-VL-32B-Instruct.}}
    \label{fusion_strength_acc}      
\end{figure}

To investigate the impact of the fusion strength $\mu$ of the V cache on the inference accuracy of VLLM, we conduct experiments on EgoSchema using the Qwen2.5-VL-32B-Instruct backbone.
Specifically, we vary the fusion strength $\mu$ from $0.1$ to $0.9$ and evaluate the corresponding inference performance.
The experimental results are shown in Figure~\ref{fusion_strength_acc} and the statistical results are summarized in Table~\ref{fusion_strength_table}.

To reduce the influence of randomness during inference, each configuration is evaluated three times independently.
We report both the mean accuracy and the variance across the three runs.
The variance reflects the stability of the VLLM under different fusion strengths.

As shown in Figure~\ref{fusion_strength_acc}, the inference accuracy reaches its maximum when the fusion strength $\mu$ is set to $0.7$.
This indicates that an appropriate fusion strength can effectively preserve informative semantic representations while reducing redundant V cache information.

When $\mu$ is smaller than $0.7$, the contribution of the fused V cache becomes insufficient, resulting in weaker preservation of useful visual information.
Consequently, important semantic features may be lost during the fusion process, leading to degraded inference accuracy.
In contrast, when $\mu$ exceeds $0.7$, the V cache information from $\mathcal{C}_{k}$ becomes excessively influenced by the V cache of $\mathcal{C}_{d}$.
Such over-fusion introduces noisy or mismatched representations into the retained cache, which contaminates the original semantic information and negatively affects the reasoning capability of the VLLM.

Another notable observation from Table~\ref{fusion_strength_table} is that the proposed method maintains relatively low variance across different fusion strengths, indicating stable inference behavior.
Although several configurations achieve comparable mean accuracy, $\mu=0.7$ not only obtains the best average performance but also maintains stable results across repeated experiments.
Therefore, we adopt $\mu=0.7$ as the default fusion strength in all remaining experiments.

\begin{table}[!h]
    \centering
    \caption{{Mean accuracy and variance under different fusion strengths $\mu$ on EgoSchema.}
    Each setting is evaluated three times independently.}
    \label{fusion_strength_table}
\resizebox{\linewidth}{!}{
\begin{tabular}{c|cc}
\toprule  
 Fusion Strength $\mu$ & Mean Accuracy & Variance \\
\midrule    
 0.1 & 59.80 & 1.15 \\
 0.2 & 60.40 & 2.30 \\
 0.3 & 60.40 & 0.03 \\
 0.4 & 58.40 & 0.44 \\
 0.5 & 58.40 & 0.75 \\
 0.6 & 59.00 & 1.00 \\
 0.7 & 65.80 & 1.04 \\
 0.8 & 59.40 & 1.31 \\
 0.9 & 59.60 & 0.25 \\
\bottomrule
\end{tabular}
}
\end{table}
\section{Compatibility with Other VLLMs}
\label{compatiable}

To further evaluate the generalization of VisCache beyond Qwen2.5-VL, we conduct experiments on two additional representative VLLM architectures: Qwen3-VL-4B-Instruct and LLaVA-OneVision~\cite{li2024llavaonevisioneasyvisualtask}. 
These VLLMs adopt different visual encoders and multimodal fusion pipelines, providing a stronger test of architectural compatibility.

\subsection{Generalization to Qwen3-VL}

For Qwen3-VL-4B-Instruct, a different-generation VLLM architecture, VisCache maintains competitive performance while reducing FLOPs by approximately 90\% compared with the full cache baseline, as shown in Table~\ref{tab:compatibility_qwen3}. Although the core ranking mechanism is backbone-agnostic, different VLLMs may exhibit different visual token distributions and varying sensitivity to visual information removal, leading to different compression--performance trade-offs. Similarly, scout models only perform temporal frame selection, and their effectiveness depends on their ability to capture task-relevant temporal cues.

We further analyze failure cases of the scout-based approach and identify three common patterns:
(1) missing transitional frames that contain important action changes;
(2) removing ambiguous frames that are necessary for contextual reasoning; and
(3) retaining redundant frames with highly similar visual content.
These cases highlight the inherent trade-off between temporal compression and information preservation, motivating our conservative retention strategy.

\begin{table}[!h]
    \centering
    \caption{
        Results of applying VisCache on the Qwen3-VL-4B-Instruct architecture.
        VisCache maintains competitive inference performance with minimal degradation compared to using full cache.
    }
    \label{tab:compatibility_qwen3}
    \resizebox{\linewidth}{!}{%
    \begin{tabular}{l|c|c|c|cc|cccc}
        \toprule
        Method & RR & FLOPs (T) & FLOPs Ratio & ActCap & DREAM1K & NExTQA & ActQA & EgoSchema & Avg \\
        Benchmark & & & & ROUGE-L & ROUGE-L & Acc & Acc & Acc & Acc \\
        \midrule
        \multicolumn{10}{c}{\textbf{Qwen3-VL-4B-Instruct}} \\
        \midrule
        Full Cache & 100\% & 5.08 & 100\% & 18.76 & 14.24 & 65.72 & 46.13 & 67.60 & 59.82 \\
        VisCache   & 28\%  & 0.52 & 10.2\% & 18.66 & 14.34 & 62.90 & 45.35 & 60.22 & 56.16 \\
        VisCache   & 19\%  & 0.51 & 10.0\% & 18.23 & 14.32 & 63.41 & 47.17 & 60.22 & 56.93 \\
        \bottomrule
    \end{tabular}%
    }
\end{table}

\subsection{Generalization to LLaVA-OneVision}

To further demonstrate the generalization capability of VisCache, we additionally evaluate our method on LLaVA-OneVision~\cite{li2024llavaonevisioneasyvisualtask}. Unlike Qwen2.5-VL, LLaVA-OneVision adopts a different visual encoder and multimodal fusion pipeline, making it a suitable benchmark for evaluating the architectural compatibility of the proposed visual KV cache compression framework. The corresponding results are presented in Table~\ref{Compatiablity_LLaVA}.

As shown in Table~\ref{Compatiablity_LLaVA}, VisCache consistently maintains competitive inference performance under aggressive visual KV cache compression. Specifically, with only $28\%$ retained visual KV cache, VisCache reduces the FLOPs from $32.73$T to $8.62$T, corresponding to only $26\%$ of the original computation cost. Despite this substantial reduction in computation, the performance degradation remains relatively limited across different benchmarks.

On ActCap, VisCache even slightly improves the ROUGE-L score compared with the full-cache setting, suggesting that removing redundant visual tokens can alleviate noisy visual representations and improve inferencing quality. Meanwhile, on DREAM1K and ActQA, VisCache preserves most of the original performance while significantly reducing the computational overhead.

\begin{table}[!h]
    \centering
    \caption{
        Results of applying VisCache on the LLaVA-OneVision architecture.
        It shows that VisCache maintains competitive inference performance across different VLLM architectures, exhibiting minimal degradation compared to using full cache.
    }
    \label{Compatiablity_LLaVA}
    \resizebox{\linewidth}{!}{%
    \begin{tabular}{l|c|c|c|ccc}
        \toprule
        Method & RR & FLOPs (T) & FLOPs Ratio & ActCap & DREAM1K & ActQA \\
        Benchmark & & & & ROUGE-L & ROUGE-L & Acc \\
        \midrule
        \multicolumn{7}{c}{\textbf{LLaVA-OneVision-Qwen2-7b-ov-hf}} \\
        \midrule
        Full Cache & 100\% & 32.73 & 100\% & 5.04 & 14.45 & 38.36 \\
        VisCache   & 28\%  & 8.62  & 26\%  & 5.52 & 11.71 & 34.97 \\
        \bottomrule
    \end{tabular}%
    }
\end{table}
Overall, these experimental results demonstrate that VisCache exhibits strong architectural compatibility and can serve as a plug-and-play visual KV cache compression framework for efficient long-video VLLM inference across different VLLM families and architectures.
The proposed coarse-to-fine visual KV cache compression strategy is not tightly coupled with a specific VLLM architecture and can generalize effectively to models with different visual encoders and fusion pipelines.


\section{Quantization}
\label{Combine with quantization}
VisCache is orthogonal to existing KV cache quantization methods and can be seamlessly combined with them for further memory reduction.
To verify this compatibility, we integrate VisCache with representative quantization methods, including KIVI~\cite{liu2024kivi} and FlatQuant~\cite{sun2024flatquant}, and evaluate the performance on multiple VQA and video understanding benchmarks.
The results are shown in Table~\ref{CombineWithQuantization}.

As shown in Table~\ref{CombineWithQuantization}, combining VisCache with 4-bit KV cache quantization still preserves competitive inference performance across different benchmarks.
On Qwen2.5-VL-3B-Instruct, VisCache combined with FlatQuant and KIVI achieves comparable or even better performance on ActQA and EgoSchema compared with the original VisCache setting.
On the larger Qwen2.5-VL-32B-Instruct backbone, although slight performance degradation is observed on several benchmarks, the overall performance remains competitive under aggressive visual KV cache compression and low-bit quantization.

These results demonstrate that VisCache is highly compatible with existing KV cache quantization frameworks and can further improve inference efficiency when combined with low-bit KV cache compression.
\begin{table*}[!h]
  \centering
  \caption{
  {Comparison of combining VisCache with different KV cache quantization methods on various VQA and VS datasets.}
  }
  \resizebox{\linewidth}{!}{
  \begin{tabular}{l|cc|cc|cccc}
    \toprule  
    Method & RR & Bits & ActCap & DREAM1K & NExtQA  & ActQA & EgoSchema & Avg. \\
    Benchmark & & & ROUGE-L & ROUGE-L & Acc  & Acc & Acc & Acc \\
    \midrule
        \multicolumn{9}{c}{\textbf{Qwen2.5-VL-3B-Instruct}} \\
    \midrule
    Full Cache & 100\% & 32 & 2.63 & 9.19 & 34.69  & 40.58 & 57.20 & 44.16 \\
    VisCache & 28\% & 32 & 2.42 & 8.70 & 41.25 &40.66 & 55.00 & 45.64 \\
    VisCache (FlatQuant) & 28\% & 4 & 2.45 & 8.58 & 40.28 & 43.07 & 56.20 & 46.52 \\
    VisCache (KIVI) & 28\% & 4 & 2.43 & 8.50 & 40.81 & 43.48 & 55.60 & 46.63 \\
    \midrule
        \multicolumn{9}{c}{\textbf{Qwen2.5-VL-32B-Instruct}} \\
    \midrule
    Full Cache & 100\% & 32 & 2.82 & 7.87 & 60.71 & 46.47 & 65.20 & 57.46 \\
    VisCache & 28\% & 32 & 2.72 & 7.29 & 62.08 & 42.69 & 65.80 & 56.86 \\
    VisCache (FlatQuant) & 28\% & 4 & 2.72 & 7.19 & 54.86 & 43.80 & 59.60 & 52.75 \\
    VisCache (KIVI) & 28\% & 4 & 2.72 & 7.09 & 54.58 & 43.88 & 58.89 & 52.45  \\    
    \bottomrule
  \end{tabular}
  }
  \label{CombineWithQuantization}
\end{table*}

\section{Comparison Under the Same Retention Ratio}
\label{app:pyramid_28}

To fairly evaluate the effectiveness of different KV cache compression strategies, we compare VisCache with several representative baselines under approximately the same retention ratio (RR). For each baseline, we tune its method-specific parameters to match the same global visual-KV retention ratio as VisCache, while keeping all other experimental conditions identical. In this section, we first report comprehensive comparisons on the Qwen2.5-VL-3B-Instruct backbone, and then provide additional results on the larger Qwen2.5-VL-32B-Instruct model.

\subsection{Results on Qwen2.5-VL-3B-Instruct}

We report results under two compression settings, i.e., approximately $28\%$ and $19\%$ retained visual KV cache, based on the Qwen2.5-VL-3B-Instruct backbone. Table~\ref{tab:same_rr_3b} summarizes the comparison across multiple video question answering and video summarization benchmarks.

As shown in Table~\ref{tab:same_rr_3b}, VisCache consistently achieves superior overall performance compared with existing methods while requiring significantly fewer FLOPs than the full-cache setting. Under the $28\%$ RR, VisCache reduces the FLOPs from $14.80$T to only $2.85$T ($19\%$ of the original computation), while still achieving the best performance on most benchmarks. In particular, VisCache obtains the highest scores on DREAM1K, ActQA, and EgoSchema, and achieves competitive performance on NExTQA. The average VQA accuracy (over NExTQA, ActQA, and EgoSchema) reaches $45.64$, outperforming the strongest baseline result of $42.92$ by $2.72$ points. Compared with methods such as PyramidKV, FastV, and PDrop, our method preserves substantially stronger video understanding capability under aggressive visual KV compression.

Under the more challenging $19\%$ RR, the advantage of VisCache becomes even more evident. Despite using only $15\%$ FLOPs of the full-cache setting, VisCache still achieves strong performance across multiple benchmarks and attains the best result on NExTQA while maintaining highly competitive results on DREAM1K and EgoSchema. The average VQA accuracy is $44.85$, exceeding the strongest baseline result of $42.53$ by $2.32$ points. In contrast, existing methods suffer from more noticeable performance degradation under the same compression level, indicating that naive KV pruning or merging strategies may discard important temporal and semantic information in long-video understanding tasks.

Another notable observation is that VisCache demonstrates significantly better robustness across different benchmarks and compression levels. While several baseline methods exhibit unstable behavior or severe accuracy drops when the RR decreases, VisCache maintains relatively stable performance. This suggests that the proposed coarse-to-fine KV compression strategy can more effectively preserve informative visual representations and reduce redundant visual tokens without severely damaging the reasoning capability of the VLLM.

\begin{table}[!t]
    \centering
    \caption{Comparison of different KV cache compression methods under the same retention ratio on Qwen2.5-VL-3B-Instruct. Bold and underlined numbers indicate the best and second-best results, respectively. Avg.\ Acc.\ denotes the average accuracy over NExTQA, ActQA, and EgoSchema.}
    \label{tab:same_rr_3b}
    \resizebox{\linewidth}{!}{
      \begin{tabular}{l|c|cc|cccc}
        \toprule  
        Method & RR & ActCap & DREAM1K & NExTQA  & ActQA & EgoSchema & Avg.\ Acc. \\
        Benchmark  &  & ROUGE-L & ROUGE-L & Acc & Acc & Acc & Acc \\
        \midrule    
        Full Cache & 100\% & 2.63 & 9.19 & 34.69 & 40.58 & 57.20 & 44.16 \\
        \midrule
        Q-Frame & 28\% & 2.37 & \textbf{9.69} & \underline{40.98} & 38.60 & 46.60 & 42.06 \\
        PyramidKV & 28\% & 2.37 & 7.34 & 38.54 & \underline{40.21} & \underline{50.00} & \underline{42.92}  \\
        FastV & 28\%  & \underline{2.43} & 8.59 & 39.60 & 39.58 & 47.60 & 42.26 \\
        PDrop & 28\% & 2.42 & 8.62 & 38.90 & 37.34 & 49.60 & 41.95 \\
        \rowcolor{blue!10}
        VisCache & 28\%  & \textbf{2.45} & \underline{8.70} & \textbf{41.25} & \textbf{40.66} & \textbf{55.00} & \textbf{45.64} \\
        \midrule
        Q-Frame & 19\% &  2.37 & 7.62 & \textbf{42.08} & 38.27 & 46.40 & 42.25 \\
        PyramidKV & 19\% & 2.36 & 7.34 & 36.38 & \textbf{40.05} & 48.50 & 41.64 \\
        FastV & 19\% & \underline{2.42} & \underline{8.13} & 38.40 & 37.20 & \underline{52.00} & \underline{42.53} \\
        PDrop & 19\% & \textbf{2.64} & 7.69 & 37.60 & 37.12 & 49.00 & 41.24 \\
        \rowcolor{blue!10}
        VisCache & 19\%  & 2.39 & \textbf{8.26} & \underline{41.44} & \underline{38.51} & \textbf{54.60} & \textbf{44.85} \\
        \bottomrule
      \end{tabular}
    }
\end{table}

\subsection{Results on Qwen2.5-VL-32B-Instruct}

To further examine whether the advantage of VisCache persists at a larger model scale, we compare it with representative baselines on the Qwen2.5-VL-32B-Instruct backbone. 
The comparison is conducted under the same $28\%$ and $19\%$ RR settings, and the results are reported in Table~\ref{tab:same_rr_32b}.

As shown in Table~\ref{tab:same_rr_32b}, VisCache maintains a clear advantage under both retention ratios. At $28\%$ RR, VisCache reduces the FLOPs to $10.75$T, corresponding to only $12\%$ of the full-cache computation, while achieving the best DREAM1K ROUGE-L of $7.29$ and the best EgoSchema accuracy of $65.80$. In comparison, Q-Frame and PDrop require $45\%$ and $22\%$ FLOPs, respectively, yet still fall behind on both metrics. At the more aggressive $19\%$ RR, VisCache uses only $10\%$ of the original FLOPs and achieves the highest EgoSchema accuracy of $65.00$, substantially outperforming Q-Frame ($56.00$) and PDrop ($57.00$). On DREAM1K, Q-Frame obtains a slightly higher ROUGE-L ($7.89$) than VisCache ($7.35$), but with $2.5\times$ more FLOPs and a much larger performance drop on EgoSchema. These results confirm that VisCache achieves a substantially better trade-off between computational efficiency and performance even on larger VLLMs.

\begin{table}[!t]
    \centering
    \caption{Comparison of representative KV cache compression methods under the same retention ratio on Qwen2.5-VL-32B-Instruct. Bold and underlined numbers indicate the best and second-best results, respectively.}
    \label{tab:same_rr_32b}
    \resizebox{\linewidth}{!}{
      \begin{tabular}{l|c|c|c|c|c}
        \toprule  
        \multirow{2}{*}{Method} & \multirow{2}{*}{RR} & FLOPs  & FLOPs & DREAM1K & EgoSchema  \\
          &  & (T) & Ratio & ROUGE-L & Acc \\
        \midrule  
        \multicolumn{6}{c}{Qwen2.5-VL-32B-Instruct} \\
        \midrule
        Full Cache & 100\% & 93.08 & 100\% &  7.87 & 65.20  \\
        \midrule
        Q-Frame & 28\% & 41.47 & 45\% &  \underline{7.16} & 62.00 \\
        PDrop & 28\% & 20.15 & 22\% &  5.16 & \underline{64.00} \\
        \rowcolor{blue!10}
        VisCache & 28\% & 10.75 & 12\% & \textbf{7.29} & \textbf{65.80}  \\
        \midrule
        Q-Frame & 19\% & 23.54 & 25\% &  \textbf{7.89} & 56.00 \\
        PDrop & 19\% & 19.51 & 21\% & 5.07 & \underline{57.00} \\
        \rowcolor{blue!10}
        VisCache & 19\% & 9.08 & 10\% & \underline{7.35} & \textbf{65.00}  \\
        \bottomrule
      \end{tabular}
    }
\end{table}

Overall, these results demonstrate that VisCache achieves a substantially better trade-off between computational efficiency and performance under matched retention budgets across different model scales, enabling efficient long-video VLLM inference under highly constrained KV cache budgets.
\section{Performance v.s. Retention Ratio}\label{app:extra_rr}
\begin{figure}[!h]
    \centering
    \includegraphics[width=0.48\textwidth]{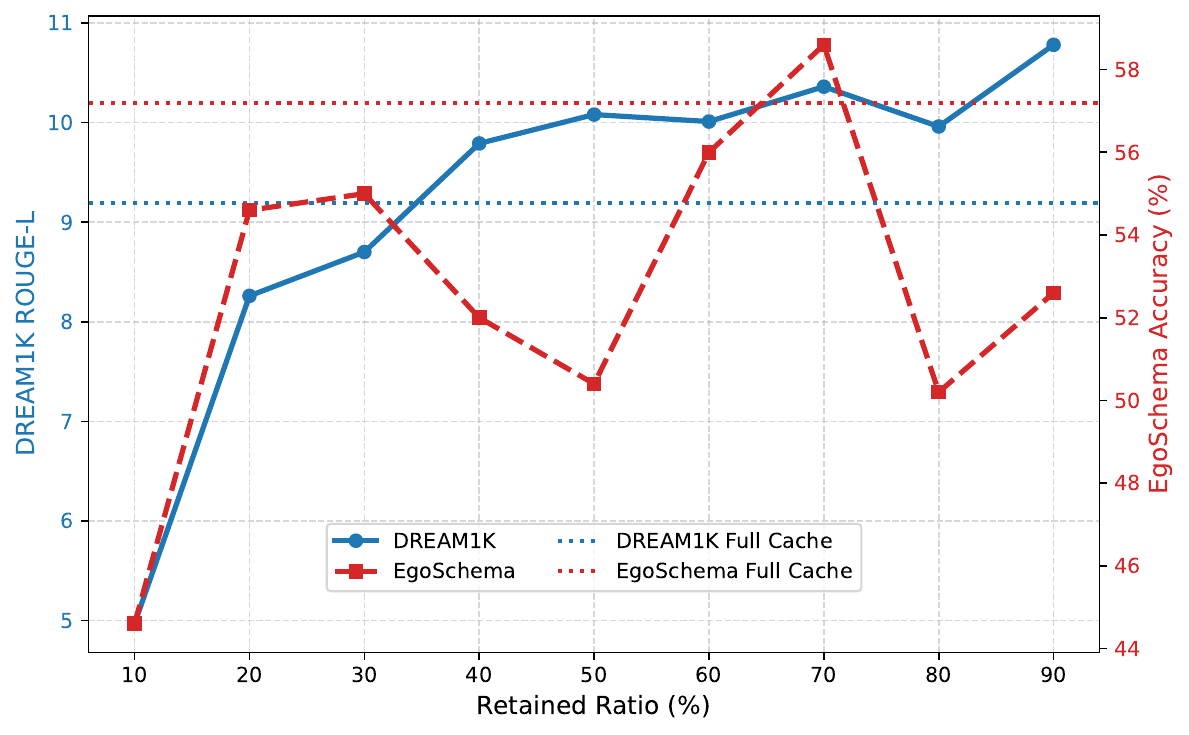}
    \caption{{Performance on DREAM1K and EgoSchema under different visual KV cache retention ratios.}
    The dashed lines indicate the full cache performance for each benchmark.}
    \label{performanceVSrr}      
\end{figure}

To evaluate the trade-off between memory efficiency and generation quality, 
we measure the performance on DREAM1K and EgoSchema under different visual KV cache RRs based on Qwen2.5-VL-3B-Instruct backbone. 
Figure~\ref{performanceVSrr} reports the results, where the dashed lines 
denote the full‑cache baselines for each benchmark.

For DREAM1K, as the retention ratio increases, the ROUGE-L score steadily improves and eventually surpasses the full cache result at around 60\% retention, reaching 10.78 at 90\%. 
This suggests that keeping more visual KV cache is beneficial for detailed description tasks, and a moderate visual KV cache budget can already outperform the unpruned setting, likely because redundant cache introduces noise.

In contrast, EgoSchema accuracy first rises and then declines after a RR of 70\%, achieving a maximum of 58.60\%. 
The drop at higher ratios indicates that preserving all visual KV cache may include distracting information for VQA, while an appropriate pruning ratio helps retain the most relevant cues.

Overall, the results highlight that optimal RRs are task‑dependent, and a well‑chosen ratio can even exceed full‑cache performance 
while reducing memory cost.

\section{Combination of baseline and Temporal Redundancy Filtering (TRF).}
\label{TRF_compatiability}
To further validate the flexibility and compatibility of the proposed Scout-based Temporal Redundancy Filtering (TRF), we integrate it with representative visual token pruning methods, including FastV and PDrop, under different retained token ratios.
The experiments are conducted on the DREAM1K and EgoSchema benchmarks based on the Qwen2.5-VL-3B-Instruct backbone.
The corresponding results are summarized in Table~\ref{tab:trf_combination}.

The results demonstrate that Scout-based TRF can be seamlessly combined with existing visual token pruning strategies and generally improves or preserves performance under aggressive token compression settings.
Specifically, under the $28\%$ RR, integrating TRF with PDrop improves the EgoSchema accuracy from $49.60$ to $52.80$, while maintaining nearly identical DREAM1K performance.
Similarly, Scout + FastV also improves EgoSchema accuracy compared with the original FastV baseline.
These results indicate that TRF can effectively complement token-level pruning strategies by removing redundant temporal information while preserving important semantic content.
\begin{table}[!t]
    \centering
    \caption{{Compatibility analysis of Scout-based Temporal Redundancy Filtering (TRF) with existing visual token pruning methods.}}
    \label{tab:trf_combination}
    \resizebox{\linewidth}{!}{
    \begin{tabular}{l|c|cc}
    \toprule  
    Method & RR & DREAM1K & EgoSchema \\
           &                & ROUGE-L & Acc \\
    \midrule    
    FastV                & 28\% & 8.59 & 47.60 \\
    PDrop                & 28\% & 8.62 & 49.60 \\
    Scout + FastV        & 28\% & 8.49 & 48.80 \\
    Scout + PDrop        & 28\% & 8.56 & 52.80 \\
    \midrule    
    FastV                & 19\% & 8.13 & 52.00 \\
    PDrop                & 19\% & 7.69 & 49.00 \\
    Scout + FastV        & 19\% & 8.42 & 49.60 \\
    Scout + PDrop        & 19\% & 8.29 & 49.40 \\
    \bottomrule
    \end{tabular}
    }
\end{table}
Under the more challenging $19\%$ RR, the effectiveness of Scout-based TRF becomes more evident on DREAM1K.
Compared with the original FastV and PDrop methods, integrating TRF improves the ROUGE-L score from $8.13$ to $8.42$ and from $7.69$ to $8.29$, respectively.
Although slight fluctuations are observed on EgoSchema, the overall performance remains competitive under highly constrained token budgets.
This suggests that Scout-based TRF can alleviate the severe information loss caused by aggressive token pruning and improve the robustness of compressed VLLM inference.

Another important observation is that the proposed TRF module is architecture-agnostic and can function as a lightweight plug-in component for existing KV cache compression or visual token pruning frameworks.
Instead of replacing previous methods, Scout-based TRF serves as a complementary temporal filtering strategy that further reduces redundant visual information across video frames.
Therefore, the proposed method exhibits strong extensibility and practical applicability for efficient long-video VLLM inference.

\section{Specific Inference Time Analysis}
\label{specific_time_analysis}
\begin{table}[!h]
    \centering
    \caption{{Detailed inference time analysis under different KV cache retention ratios.}
    We report the end-to-end (E2E) latency, time to first token (TTFT), and time per output token (TPOT) on DREAM1K and EgoSchema benchmarks based on Qwen2.5-VL-3B-Instruct backbone. 
    Our VisCache consistently reduces inference latency as the KV cache RR decreases, achieving up to 2.35$\times$ E2E speedup while maintaining efficient generation performance.}
\resizebox{\linewidth}{!}{
\begin{tabular}{l|c|c|c|c}
\toprule
Method  & RR & E2E Latency (s) & TTFT (s) & TPOT (ms)  \\
\midrule
\multicolumn{5}{c}{\textbf{DREAM1K}}\\
\midrule
Full Cache & 100\% & 12.24 & 4.42 & 118.81   \\
VisCache & 40\% & 9.55 (1.28$\times$) & 2.05 & 119.06 \\
VisCache   & 28\% & 6.33 (1.93$\times$) & 3.45 & 45.79   \\
VisCache   & 19\% & 5.20 (2.35$\times$) & 2.23 & 45.60  \\
\midrule
\multicolumn{5}{c}{\textbf{EgoSchema}}\\
\midrule
Full Cache & 100\% & 13.92 & 13.50 & 59.37 \\
VisCache & 40\% & 10.47 (1.33$\times$) & 10.08 & 56.82 \\
VisCache   & 28\% & 9.32 (1.49$\times$) & 8.97 & 49.50  \\
VisCache   & 19\% & 8.30 (1.68$\times$) & 7.92 & 47.70 \\
\bottomrule
\end{tabular}
}
\label{tab:time_analysis}
\end{table}
Table~\ref{tab:time_analysis} presents a detailed latency breakdown of VisCache under different visual KV cache RR.  
Across both DREAM1K and EgoSchema benchmarks, reducing the KV cache RR consistently decreases overall inference latency, demonstrating the effectiveness of VisCache in accelerating VLLM inference.

Specifically, on DREAM1K, VisCache achieves up to 2.35$\times$ E2E speedup at 19\% RR, while TPOT is reduced from 118.81 ms to 45.60 ms. 
Similarly, on EgoSchema, VisCache reduces E2E latency from 13.92 s to 8.30 s and improves decoding efficiency by lowering TPOT from 59.37 ms to 47.70 ms. 

Overall, the results demonstrate that VisCache achieves an effective balance between visual KV cache compression and practical system efficiency, providing substantial inference acceleration while maintaining stable decoding performance.

\section{Impact of the Scout VLM}
\label{impactvlm}
\begin{table}[!t]
    \centering
    \caption{
        Effect of different scout VLMs for temporal filtering.
        We keep $\text{RR}=28\%$ of the visual KV cache on Qwen2.5-VL-32B-Instruct across all variants.
    }
    \label{tab:scout_ablation}
    \resizebox{\linewidth}{!}{
        \begin{tabular}{l|c|cc|ccc}
        \toprule
        Method & Scout & ActCap & DREAM1K & NExTQA & ActQA & EgoSchema \\
        & Model & ROUGE-L & ROUGE-L & Acc & Acc & Acc \\
        \midrule
        Full Cache & -- & 2.63 & 9.19 & 34.69 & 40.58 & 57.20 \\
        \midrule
        \multirow{3}{*}{VisCache} & CLIP      & 2.42 & 8.70 & 41.25 & 40.66 & 55.00 \\
        & BLIP      & 2.44 & 8.49 & 37.18 & 40.81 & 46.80 \\
        & OpenCLIP  & 2.43 & 8.51 & 41.56 & 39.78 & 53.40 \\
        \bottomrule
        \end{tabular}
    }
\end{table}
We further investigate how the choice of the scout model affects overall performance.
Table~\ref{tab:scout_ablation} compares CLIP~\cite{radford2021learning}, BLIP~\cite{li2022blip}, and OpenCLIP~\cite{cherti2023reproducible} as the frame-filtering scout under identical VisCache settings.
The results reveal that the scout choice has a non-trivial impact on downstream VQA accuracy, with CLIP and OpenCLIP notably outperforming BLIP on several benchmarks, while all three scouts perform comparably on the VS tasks.
This suggests that the semantic alignment between the scout's visual encoder and the main LLM backbone plays an important role in selecting frames that facilitate accurate question answering.
We adopt CLIP as the default scout for its balanced performance across both task types.
Notably, even with the least effective scout, VisCache retains a substantial portion of full-cache accuracy while reducing the KV cache to 28\%, confirming that the dual-stage framework is robust to the choice of the pre-filtering model.
The rationale and compatibility of the scout VLM are further discussed in Appendix~\ref{The Effect of Selecting Keyframes by Small VLM} and Appendix~\ref{TRF_compatiability}.

\section{Inference Example}
\label{Examples}

\begin{figure*}[t]
    \centering
    \includegraphics[width=0.96\textwidth]{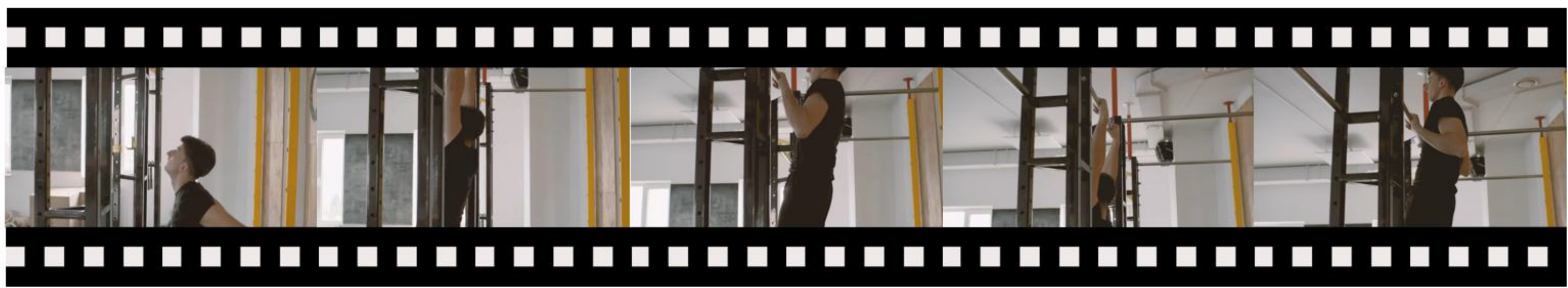}
    \caption{Comparison between Full Cache and VisCache on a pull-up activity video.}
    \label{fig:pullup_example}
\end{figure*}
\begin{figure*}[t]
    \centering
    \includegraphics[width=0.96\textwidth]{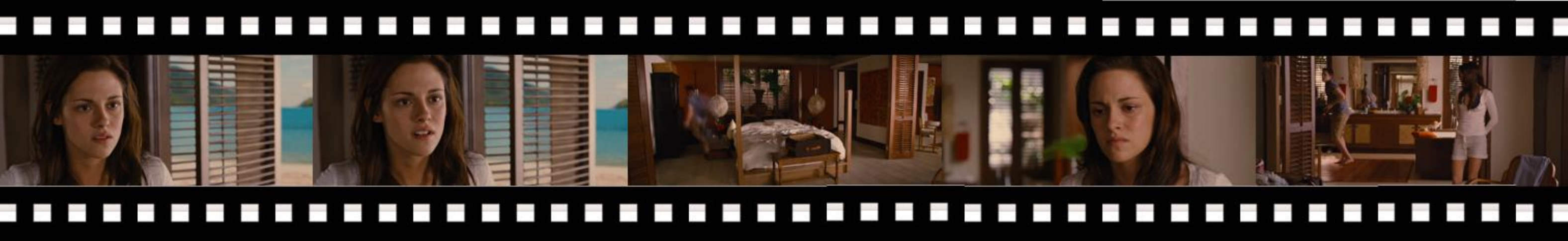}
    \caption{Comparison between Full Cache and VisCache on a movie scene video.}
    \label{fig:moon_example}
\end{figure*}
\noindent\textbf{User:} Describe the content of this video in detail.

\noindent\textbf{Full Cache:}
A young man is performing pull-ups on a bar in a gym. He is wearing a black T-shirt and black shorts. The gym has white walls and a ceiling with exposed pipes. The man starts by hanging from the bar with his arms fully extended, then pulls himself up until his chin is over the bar.

\noindent\textbf{VisCache:}
The video begins with a man standing in front of a door, preparing to perform a pull-up exercise. He is wearing a black tank top and shorts, and his hair is tied back. The setting appears to be an indoor gym or workout area, with a metal pull-up bar mounted on a wall.


\noindent\textbf{User:} Describe the content of this video in detail.

\noindent\textbf{Full Cache:}
The video begins with a close-up shot of a woman with long brown hair, looking off to the side with a serious expression. The scene then cuts to a man with short dark hair, wearing a blue shirt and a black vest, who is looking at something off-screen with a concerned expression. The camera then cuts to a woman in a white shirt and shorts walking through a room, her back facing the camera. She turns around and walks towards a bed, where she bends over and picks up a suitcase. The camera then cuts to a close-up of a woman's face, who appears to be crying.

\noindent\textbf{VisCache:}
The video begins with a close-up shot of a woman's face, showing her looking out of a window. The scene then transitions to a man walking through a doorway into a room. He is wearing a white shirt and appears to be in motion. The camera follows him as he moves through the room, which has a modern design with a large window that offers a view of the outdoors. The man continues to walk towards the camera, and the video ends with him standing in front of it.

\end{document}